\documentclass[journal]{IEEEtran}

\usepackage{cite}
\usepackage{orcidlink}
\usepackage{hyperref}
\usepackage{makeidx}
\usepackage{url}
\usepackage{graphicx}
\usepackage{enumitem}
\usepackage{amsmath}
\usepackage{amsfonts}
\usepackage{amssymb}
\usepackage{amsthm}
\usepackage{algorithmic}
\usepackage{booktabs}
\usepackage{tabularx}
\usepackage{makecell}
\usepackage{float}
\usepackage{mathtools}
\usepackage{caption}
\usepackage{subcaption}
\usepackage{adjustbox}
\usepackage{multirow}

\hypersetup{hidelinks, colorlinks=true, citecolor=blue, linkcolor=blue, urlcolor=black}

\DeclareMathOperator{\tr}{tr}

\theoremstyle{definition}
\newtheorem{assumption}{Assumption}
\newtheorem{definition}{Definition}
\newtheorem{lemma}{Lemma}
\newtheorem{proposition}{Proposition}

\newtheorem{remark}{Remark}
\newtheorem{corollary}{Corollary}

\usepackage[ruled,vlined,linesnumbered]{algorithm2e}
\SetAlCapNameFnt{\footnotesize}
\SetAlCapFnt{\footnotesize}
\SetAlgorithmName{Algorithm}{Algorithm}{List of Algorithms}

\newcommand{\refig}[1]{Fig.~\ref{#1}}
\newcommand{\refalg}[1]{Alg.~\ref{#1}}

\newcommand{\refsec}[1]{Sec.~\ref{#1}}

\begin{document}

\title{R-SFLLM: Jamming Resilient Framework for Split Federated Learning with Large Language Models}

\author{Aladin Djuhera \orcidlink{0009-0005-1641-8801},~\IEEEmembership{Graduate Student Member,~IEEE,}
        Vlad~C.~Andrei \orcidlink{0000-0001-5443-0100},~\IEEEmembership{Graduate Student Member,~IEEE,} \\
        Xinyang~Li \orcidlink{0000-0001-7262-5948},~\IEEEmembership{Graduate Student Member,~IEEE,}
        Ullrich~J.~M\"onich \orcidlink{0000-0002-2390-7524},~\IEEEmembership{Senior Member,~IEEE,} \\
        Holger~Boche \orcidlink{0000-0002-8375-8946},~\IEEEmembership{Fellow,~IEEE,}
        and~Walid~Saad \orcidlink{0000-0003-2247-2458},~\IEEEmembership{Fellow,~IEEE}

\thanks{
\noindent
The authors were supported by the German BMBF under the program “Souver\"an. Digital. Vernetzt.” as part of the research hub 6G-life (Grant 16KISK002), QD-CamNetz (Grant 16KISQ077), QuaPhySI (Grant 16KIS1598K), and QUIET (Grant 16KISQ093). W. Saad was supported by the U.S. National Science Foundation under Grant CNS-2114267.
}
}

\maketitle

%
\IEEEpeerreviewmaketitle

\begin{abstract}
    \noindent
    Split federated learning (SFL) is a compute-efficient paradigm in distributed machine learning (ML), where components of large ML models are outsourced to remote servers. 
    A significant challenge in SFL, particularly when deployed over wireless channels, is the susceptibility of transmitted model parameters to adversarial jamming that could jeopardize the learning process. 
    This is particularly pronounced for embedding parameters in large language models (LLMs) and vision language models (VLMs), which are learned feature vectors essential for domain understanding.
    In this paper, rigorous insights are provided into the influence of jamming embeddings in SFL by deriving an expression for the ML training loss divergence and showing that it is upper-bounded by the mean squared error (MSE). 
    Based on this analysis, a physical layer framework is developed for resilient SFL with LLMs (R-SFLLM\footnote{\label{repo} Code available at: \url{https://github.com/aladinD/R_SFLLM}}) over wireless networks. R-SFLLM leverages wireless sensing data to gather information on the jamming directions-of-arrival (DoAs) for the purpose of devising a novel, sensing-assisted anti-jamming strategy while jointly optimizing beamforming, user scheduling, and resource allocation. 
    Extensive experiments using both LLMs and VLMs demonstrate R-SFLLM's effectiveness, achieving close-to-baseline performance across various natural language processing (NLP) and computer vision (CV) tasks, datasets, and modalities. 
    The proposed methodology further introduces an adversarial training component, where controlled noise exposure significantly enhances the model's resilience to perturbed parameters during training. 
    The results show that more noise-sensitive models, such as RoBERTa, benefit from this feature, especially when resource allocation is unfair. 
    It is also shown that worst-case jamming in particular translates into worst-case model outcomes, thereby necessitating the need for jamming-resilient SFL protocols.
\end{abstract}


\section{Introduction and Motivation}
\label{sec:introduction}
Future 6G networks are anticipated to introduce a substantial leap toward highly integrated and intelligent connectivity at the edge, enabled by artificial intelligence (AI) \cite{saad2024artificial} and machine learning (ML) \cite{saad2019vision}. 
However, in this envisioned hyper-connected and AI-assisted network, important questions and stringent requirements on resilience and trustworthiness arise, both from a user and network perspective \cite{sec_trust_6g}. 
This imposes significant design challenges for emerging technologies, such as distributed and collaborative ML (DCML) \cite{chen2021distributed}, which may be targeted by adversarial attacks over the wireless medium. 
For instance, the distributed training of large language models (LLMs) faces unique challenges in ensuring data integrity and model resilience due to highly sensitive embedding parameters. 
Much research has focused on addressing these challenges via adversarial AI training methods \cite{TIFS_paper} and model-based solutions \cite{qiu2019review}, and in \cite{son2024adversarial}, the authors closely investigated such attack and defense strategies, particularly in 6G networks, exposing critical vulnerabilities across all network layers. 
However, a number of important research questions remain underexplored, such as how these attacks can be orchestrated in practice, and how current and next-generation wireless network architectures, systems, and technologies can introduce \emph{proactive} defense mechanisms, ideally by-design.  

\vspace{-0.1cm}

\subsection{Adversarial Poisoning in Wireless Federated LLM Training.}
Motivated by the increasing importance of end-user data privacy and associated privacy protection laws \cite{ccpa2018}, DCML has been gradually shifting toward mechanisms such as federated learning (FL) \cite{mcmahan2023communicationefficient} and split FL (SFL) \cite{thapa2022splitfed}. 
These have proven to be privacy-preserving as only the respective model parameters, some parts of them, or model gradients need to be exchanged. SFL in particular has emerged as a compute-efficient federated protocol, suitable for distributed training of large ML model architectures. 
Unlike traditional FL, in which the entire model is trained on each client, SFL splits the model, allowing for more compute-intensive parts to be outsourced to a remote server. 
This approach is particularly advantageous for LLM architectures that cannot be entirely processed at resource-constrained edge devices due to computational and memory limitations \cite{jiao2020tinybert}. 
However, the practical realization of (S)FL over wireless networks faces challenges from the inherently unreliable wireless medium, limited bandwidth, and suboptimal resource allocation \cite{chen2020wireless}. 
In addition, malicious actors such as jammers could target privacy and security aspects of (S)FL systems by intentionally poisoning data and models through adversarial noise. 
The particular importance of studying adversarial jamming attacks in SFL with LLMs is motivated by recent results from natural language processing (NLP) research \cite{yang2021careful, yoo2022backdoor}. 
The authors in \cite{yang2021careful} study the susceptibility of LLMs to word embedding poisoning caused by noisy perturbations and show that by altering even a single embedding vector, an adversary can subtly manipulate a model to react abnormally to specific trigger words. 
Moreover, the severity of such embedding poisoning attacks for federated networks was studied in \cite{yoo2022backdoor}, showing that even a small number of compromised clients is sufficient to effectively deteriorate the global model. 
In federated systems over wireless networks, adversarial jamming emerges as a realistic threat for orchestrating such poisoning attacks by corrupting sensitive word or other multi-modal embeddings during transmission. 
This covert adversarial perspective is in contrast to prior works on jamming in FL \cite{shi2022jamming, anti_jamming_medicalIoT}, which do not consider the implications of poisoning the model's reasoning capabilities. 
This is particularly critical in SFL, where embeddings might be directly transmitted as intermediate split parameters. 
However, the detrimental impact of such attacks in SFL remains underexplored.  

\subsection{Proactive and Resilient-by-Design Anti-Jamming in SFL.}
To preemptively safeguard LLM parameter transmissions in SFL against adversarial jamming attacks, \emph{proactive} and in particular \textit{resilient-by-design} approaches are needed \cite{fettweis_boche_resilience2022}. 
This requires a simultaneous co-design of AI, resilience, beamforming, user scheduling, and resource allocation, thereby integrating resilience from a bottom-up approach, by design.
In addition, such proactive defense mechanisms need to be agnostic of the respective jamming capabilities, including physical and spatial features. 
This is not the case for some more recent works, which tend to impose strong assumptions on the adversary's knowledge and setup. 
For instance, \cite{marti_mitigating_2023} and \cite{marti_universal_2023} only consider single- or few-antenna jammers with common secrets being exchanged between legitimate parties. 
This essentially excludes so-called \textit{worst-case} jammers with extensive system knowledge and capabilities \cite{worst_case_jamming_mimo, corr_jamming}. 
To develop universally applicable defense strategies for a wide range of jamming scenarios in SFL, system performance needs to be guaranteed for the worst-case. 
Thus, we need to generalize toward intelligent and reconfigurable worst-case jammers. 
In our prior work in \cite{andrei2024resilientbydesign, 10901454}, we studied how \emph{sensing-assisted} network information can be harnessed to enhance existing mitigation schemes without the need for otherwise precise jamming statistics. 
Therein, we have shown that information on the jamming signal directions-of-arrival (DoAs) can be used to devise MIMO-OFDM anti-jamming strategies with exceptional performance. 
However, we did not discuss whether such sensing-assisted defense strategies can be straightforwardly applied to enhance the resilience in SFL over wireless networks. 
In particular, the impact of worst-case jamming needs to be quantified in order to study its influence on LLM model poisoning as compared to conventional jammers. 
In the subsequent sections, we provide thorough insights on these aspects, including an analysis on the minimum system rate that guarantees a reliable and resilient SFL training.

\subsection{Contributions.}
The main contribution of this paper is an analysis and framework for resilient SFL over wireless networks that will help close the gap between adversarial jamming attacks in SFL and LLM model poisoning. 
Here, \emph{resilience} transcends conventional \emph{robustness} by \emph{proactively} integrating physical-layer defenses, ensuring the global training process remains effective even under worst-case jamming, rather than merely withstanding moderate perturbations.
We provide insights into how jamming of LLM embeddings affects the global model training and how the latter can be effectively safeguarded by MIMO signal processing at the availability of sensing-assisted DoA information. 
In summary, our key contributions include:
\begin{enumerate}[leftmargin=*]
    \item \textbf{Analytical Bound for LLM Loss Divergence:} We derive a novel expression for the training loss divergence in case of corrupted embeddings under a relaxed $(\boldsymbol{L}_0, \boldsymbol{L}_1)$-smoothness assumption, and show that its upper bound depends on the communication mean squared error (MSE), thereby motivating a \emph{wireless} approach to resilience in SFL.

    \item \textbf{Minimum System Rate Guarantees for SFL:} We provide a novel analysis on the minimum system rate that ensures reliable SFL over wireless networks. By relating outage rates to jamming power and user scheduling, we characterize key \emph{scalability} conditions and illustrate how larger-scale SFL remains feasible under adversarial interference.

    \item \textbf{Sensing-Assisted Anti-Jamming Framework:} We develop R-SFLLM, a novel, sensing-assisted anti-jamming framework for resilient SFL with LLMs, which leverages the jammer’s DoAs to devise an anti-jamming strategy formulated as a joint optimization problem for beamforming, user scheduling, and resource allocation while maximizing the sum rate of the SFL participants. 
    In this problem, any explicit knowledge about the jamming statistics is replaced by a surrogate expression that depends only on the jamming DoAs. 
    We provide an efficient solution to the problem using an iterative water-filling approach \cite{yu2004iterative}, thereby going beyond mere algorithmic robustness and enabling \emph{resilience-by-design} through the physical layer.

    \item \textbf{Worst-Case Jamming:} To validate R-SFLLM under severe threat conditions, we develop the \emph{worst-case} jamming strategy \cite{andrei2024resilientbydesign}, which minimizes the total system sum rate.
    
    \item \textbf{Extensive Experiments:} We provide results for NLP-based BERT \cite{devlin2018bert} and RoBERTa \cite{liu2019roberta} LLMs, as well as computer vision (CV)-based CLIP \cite{CLIP} vision language models (VLMs), covering various tasks across 13 datasets and two modalities. 
    We demonstrate \emph{near-optimal} performance when anti-jamming is enabled and significantly worse outcomes for unprotected scenarios. 
    Further, we show that R-SFLLM introduces an \emph{adversarial training} component as embeddings are exposed to controlled noise since jamming cannot be mitigated perfectly, thus improving the model's resilience by teaching it to learn effectively in the presence of interference \cite{jain2023neftune}.
    We also provide ablation studies on scalability and jamming capabilities.
\end{enumerate} 

The rest of this paper is organized as follows. 
Section II presents the R-SFLLM system model and derives expressions for the ML loss divergence and the minimum system rate.
In Section III, we present the anti-jamming framework and develop the worst-case jamming strategy. 
Section IV discusses the simulation results and Section V concludes the paper.

\section{System Model and Adversarial Analysis}
\label{sec:system_model_and_analysis}

\subsection{Wireless R-SFLLM System Model.}
\label{sec:system_model}
\begin{figure}[t]
    \centering
    \includegraphics[width=0.49\textwidth]{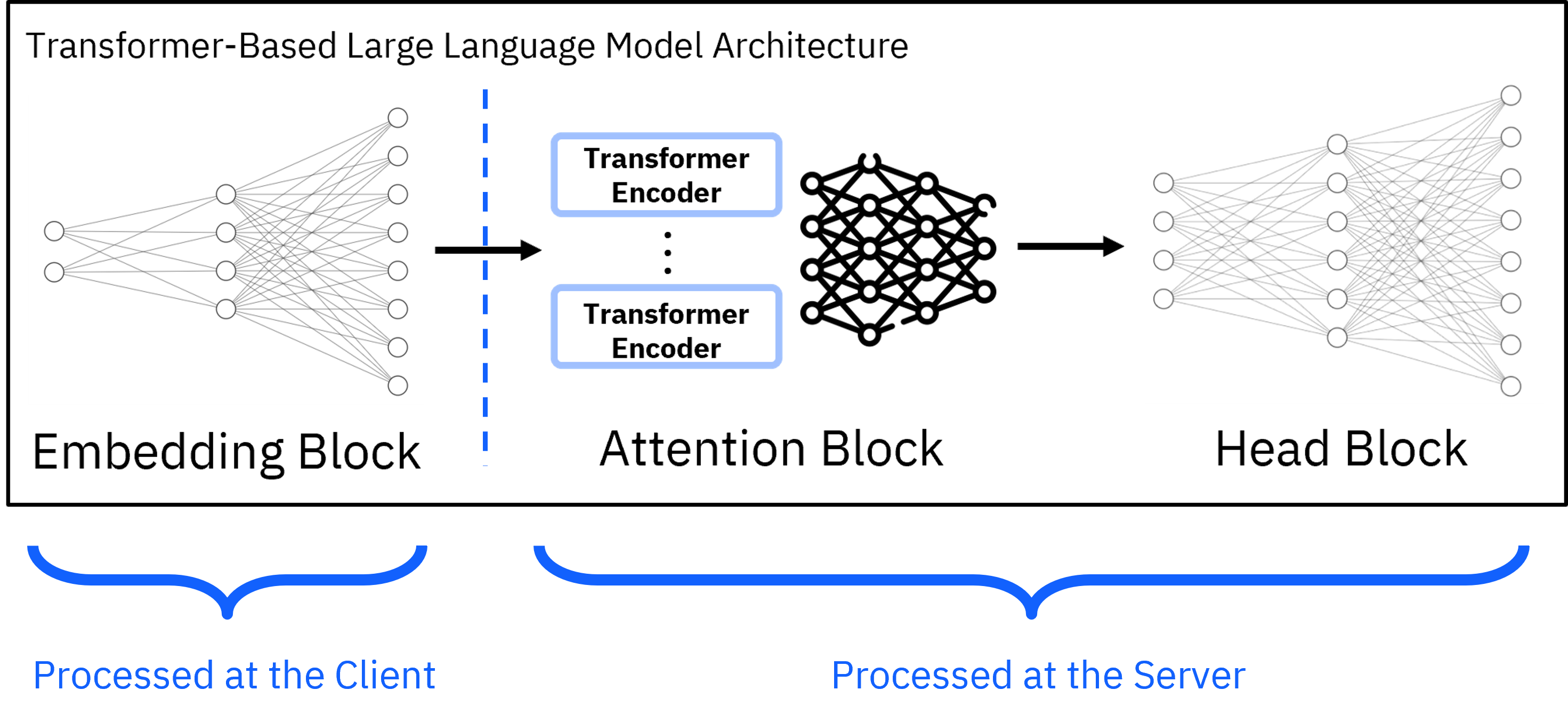}
    \vspace{-0.5cm}
    \caption{SFL model split with LLM embeddings being processed at the client and with attention and head layers being processed at the server.}
    \vspace{-0.5cm}
    \label{fig:LLM_split}
\end{figure}

We consider an SFL setup in which a set $\mathcal{Q}$ of $Q$ legitimate clients cooperatively train transformer-based LLMs (analogously for VLMs), which consist of embedding, attention, and head layers. 
A natural choice in SFL is to partition the model according to these blocks, assigning the embedding layer to the client and the compute-intensive attention and head layers to the server, as shown in \refig{fig:LLM_split}. 
This particular partitioning alleviates the computational load at the client while ensuring that embeddings are processed close to the raw data, thereby enhancing privacy. 
Further partitioning the embedding block and transmitting intermediate layers instead increases the risk of sensitive information being exposed to adversarial attacks, such as model inversion \cite{geiping2020inverting}. 
During training, each user $q \in \mathcal{Q}$ first computes the embeddings $\boldsymbol{e}_q \in \mathbb{R}^{E}$ for its private data points and then maps them onto uncorrelated zero-mean, unit variance Gaussian symbols, which are then beamformed, power-scaled, and transmitted over the wireless channel to a dedicated server slice for further processing. 
To this end, we consider a MIMO-OFDM multiple access channel (MAC) in the uplink. 
Each user $q$ is equipped with $N_{T_q}$ antennas and transmits the signal $\boldsymbol{x}_{qnk}$, which is a composite of the binary user scheduling $\alpha_{qnk}$, transmit power $p_{qnk}$, beamforming vector $\boldsymbol{w}_{qnk} \in \mathbb{C}^{N_{T_q}}$, and embedding data symbols $s_{qnk}$. 
The transmissions occur over the resource set $\mathcal{R}_q = \mathcal{N}_q \times \mathcal{K}_q$ with allocated subcarriers $n \in \mathcal{N}_q$ and OFDM symbols $k \in \mathcal{K}_q$, with a total of $N$ subcarriers and $K$ symbols available. 
Each legitimate transmit signal propagates through the channel $\boldsymbol{H}_{qnk} \in \mathbb{C}^{N_R \times N_{T_q}}$ to the server with $N_{R}$ receive antennas. 
An adversarial jammer aims to impair the SFL training by jamming the embeddings in the uplink. 
The legitimate signal is thus corrupted by additive white Gaussian noise (AWGN) $\boldsymbol{\eta}_{nk} \sim \mathcal{N}(\boldsymbol{0}, \sigma^2 \boldsymbol{I}) \in \mathbb{C}^{N_R}$ and by the adversarial jamming signal $\boldsymbol{u}_{nk} \sim \mathcal{N}(\boldsymbol{0}, \boldsymbol{C}_{\boldsymbol{u}_{nk}}) \in \mathbb{C}^{N_J}$, which propagates through the separate jamming channel $\boldsymbol{G}_{nk} \in \mathbb{C}^{N_R \times N_J}$. 
We define $N_J$ as the number of jamming antennas and corresponding jamming covariance matrix as $\boldsymbol{C}_{\boldsymbol{u}_{nk}} \in \mathbb{C}^{N_J \times N_J}$. 
The receiver performs equalization using the linear filters $\boldsymbol{v}_{qnk}^H \in \mathbb{C}^{1 \times N_R}$ to estimate the transmitted symbols $\hat{s}_{qnk}$. 
In summary, we have: 
\begin{align} 
    &\boldsymbol{x}_{qnk} = \alpha_{qnk} \cdot \sqrt{p_{qnk}} \boldsymbol{w}_{qnk} s_{qnk} \in \mathbb{C}^{N_{T_q}} \label{eq:x_qnk}, \\
    &\boldsymbol{z}_{nk} = \boldsymbol{G}_{nk} \boldsymbol{u}_{nk} + \boldsymbol{\eta}_{nk} \in \mathbb{C}^{N_R}, \\
    &\boldsymbol{y}_{nk} = \sum_{q \in \mathcal{Q}} \boldsymbol{H}_{qnk} \boldsymbol{x}_{qnk} + \boldsymbol{z}_{qnk} \in \mathbb{C}^{N_R}, \\
    &\hat{s}_{qnk} = \boldsymbol{v}_{qnk}^H \boldsymbol{y}_{nk} \label{eq:symbol_estimate}.
\end{align}

We further model $\boldsymbol{H}_{qnk}$ and $\boldsymbol{G}_{nk}$ as beamspace channels:
\begin{align}
    &\boldsymbol{H}_{qnk} = \sum_{l=1}^{L_{H_q}} b_{H_{q,l}} \boldsymbol{a}_{N_R}(\boldsymbol{\theta}_{q, l}) \boldsymbol{a}_{N_{T_q}}^H (\boldsymbol{\psi}_{q,l})  e^{j 2\pi \omega_{nk} (\nu_{q,l} , \tau_{q,l})} \label{eq:beamspace_h} \\
    &\boldsymbol{G}_{nk} = \sum_{l=1}^{L_{G}} b_{G,l} \boldsymbol{a}_{N_R}(\boldsymbol{\theta}_{G, l}) \boldsymbol{a}_{N_J}^H (\boldsymbol{\psi}_{G,l}) e^{j 2\pi \omega_{nk} (\nu_{G,l} , \tau_{G,l})}. \label{eq:beamspace_g}
\end{align}

\begin{figure*}[htbp]
    \centering
    \includegraphics[width=1\textwidth]{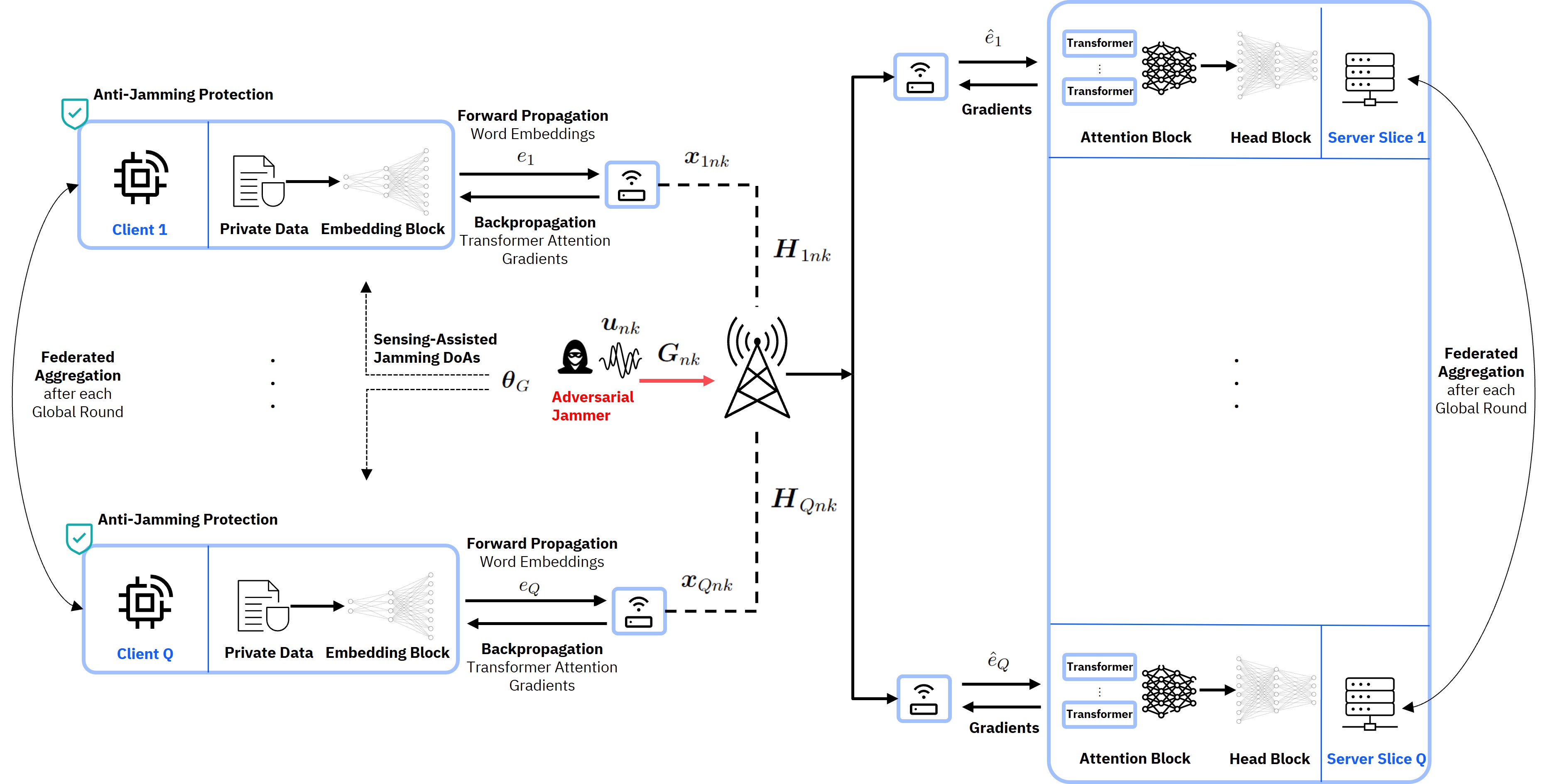}
    \caption{R-SFLLM system architecture for distributed training over MIMO-OFDM wireless channels, augmented by sensing-assisted anti-jamming capabilities.}
    \label{fig:sfl_simulation_setup}
\end{figure*}

Here, $L_{H_q}$ and $L_G$ are the number of resolvable paths $l$ for each channel, $b_{\boldsymbol{\cdot},l}$ is the path gain, $\boldsymbol{a}_{N_X}(\boldsymbol{\theta})$ is the steering vector at each terminal with $N_X$ antennas, $\boldsymbol{\theta}_{\boldsymbol{\cdot},l}$ is direction-of-arrival, $\boldsymbol{\psi}_{\boldsymbol{\cdot},l}$ is direction-of-departure, and $w_{nk} (\nu, \tau) = k \nu T_s - n \tau \Delta f$ is the phase shift caused by the Doppler shift $\nu$ and propagation delay $\tau$, with $T_s$ and $\Delta f$ being the symbol period and subcarrier spacing. 
Furthermore, the power $P_q$ for each user is limited across all resource elements, and the jamming signal equally adheres to a jamming power constraint $P_J$, i.e.
\begin{align}
    \sum_{(n,k) \in \mathcal{R}_q} 	\| \boldsymbol{x}_{qnk} \|_2^2 \leq P_q, \
    \sum_{(n,k) \in \mathcal{R}_q \forall q} \tr(\boldsymbol{C}_{\boldsymbol{u}_{nk}}) \leq P_J.
\end{align}

We further assume that the legitimate parties have precise channel state information (CSI) (e.g., via pilot‑echoes), encompassing the wireless link parameters defined by the set $\zeta_q = \{\alpha_{qnk}, p_{qnk}, \boldsymbol{w}_{qnk}, \boldsymbol{v}_{qnk}, \boldsymbol{H}_{qnk}, \sigma^2\}$. 
Additionally, the SFL participants are provided with the DoAs of the jamming signal, i.e. $\boldsymbol{\theta}_G = \{\theta_{G,l}\}_{l=1}^{L_G}$. 
This may be enabled by advanced wireless sensing technologies in future 6G networks, such as integrated sensing and communication (ISAC) and reflective intelligent surfaces (RIS) \cite{ris1, ris2, chaccour2024joint, zhu2024enabling, gonzalez2024integrated}, to name a few, thus making it a realistic assumption in practice.
Further, we assume no restrictions on the particular jamming strategy, hence the jammer is assumed to be in the so-called \textit{jammer-dominant regime} \cite{worst_case_jamming_mimo} with more transmit power and antennas than any legitimate party, i.e. $P_J \gg P_q$ and $N_J > N_{T_q}, N_{R}$. 
In addition, the adversary may possess full system knowledge, including $\zeta_q$. 
This represents a \emph{worst-case} jammer assumption. 

Adversarial jamming introduces corruption not only at the symbol level but also at the decoded message, such that it can be modeled as the post-decoding error as follows: 
\begin{align}
    \hat{\boldsymbol{e}}_q &= \boldsymbol{e}_q + \epsilon, \ \text{where} \ \epsilon \sim \mathcal{N} \left ( 0, \boldsymbol{C}_{\epsilon} \right ) \label{eq:noise_model}, \\
    \tr ( \boldsymbol{C}_{\epsilon} ) &= \text{MSE}(\boldsymbol{s}_q), \\
    \text{MSE}(\boldsymbol{s}_q) &= \mathbb{E} \left [ \| \boldsymbol{s}_q - \hat{\boldsymbol{s}}_q \|_2^2 \right ] \nonumber \\ 
    &= \sum_{(n,k) \in \mathcal{R}_q} \alpha_{qnk} \cdot \mathbb{E} \left [ | s_{qnk} - \hat{s}_{qnk} |^2 \right ].
\end{align}

Upon receiving the jammed signal $\boldsymbol{y}_{nk}$, the server continues processing the LLM attention and head layers using the corrupted embeddings $\hat{\boldsymbol{e}}_q \neq \boldsymbol{e}_q$. 
At the end of the forward propagation pass, the server computes the training loss $L: \mathbb{R}^E \rightarrow \mathbb{R}$, which yields the corrupted loss $L(\hat{\boldsymbol{e}}_q)$ and its gradient $\nabla L(\hat{\boldsymbol{e}}_q)$, using which the backpropagation process is initiated. 
This procedure is repeated for each transmission of the embeddings across all global training rounds. We also assume that the jammer is not active in the downlink as the perturbation of gradients has been studied in various federated setups, for which corresponding defense mechanisms exist \cite{wei2021gradient}. 
Similarly, the client- and server-side model aggregation after each global round are assumed to be unaffected by the adversary as corresponding secure aggregation strategies exist as well \cite{bonawitz2016practical}. 
Note that FedAvg \cite{mcmahan2023communicationefficient} is used in this work. 
Thus, the consideration of only the uplink transmission suffices to study the impact of jamming LLM embeddings in this setup. 
Anti-jamming can then be directly applied if necessary conditions are fulfilled, including the assumption that maximizing the signal-to-interference-plus-noise-ratio (SINR) implies maximizing the ML performance. 
This assumption is verified next. 
\refig{fig:sfl_simulation_setup} shows the R-SFLLM system architecture, where the SFL protocol is augmented by sensing-assisted jamming DoA information, a necessary component for our anti-jamming framework presented in Section \ref{sec:r_sfllm_framework}.
\vspace{-2em}
\subsection{Adversarial Jamming Impact on LLM Training in SFL.}
\label{sec:jamming_impact}
Previous works in (S)FL typically assume the loss function to be \textit{convex}, \textit{twice differentiable}, and \textit{Lipschitz smooth}.
While these assumptions may hold true for simpler neural networks as in \cite{chen2020wireless} and \cite{chenjoint}, more involved architectures such as transfomers generally do not exhibit these properties \cite{robustness}. 
The assumption that $L$ is Lipschitz smooth is particularly far-reaching as this implies \textit{bounded gradients} during backpropagation. 
In \cite{zhang2019gradient}, it is shown that the standard Lipschitz assumption introduces a large variability along the optimization trajectory. 
Thus, a relaxed $(\boldsymbol{L_0}, \boldsymbol{L_1})$-smoothness needs to be assumed, which generalizes to more complex models, such as LLMs/VLMs. 
Based on this generalization, we derive upper bounds on the loss divergence, caused by jammed embeddings, and show how these relate to the communication MSE.

\subsubsection{Assumptions on the Loss Function}

\begin{assumption}\label{A1}
    The loss function $L: \mathbb{R}^{E} \rightarrow \mathbb{R}$ is twice-differentiable and bounded from below with infimum $L^*$.
\end{assumption}

\begin{assumption}\label{A2}
    $L$ is $(\boldsymbol{L_0}, \boldsymbol{L_1})$-smooth coordinate-wisely, i.e. there exist coefficient vectors $\boldsymbol{L_0}, \boldsymbol{L_1} \in \mathbb{R}^E$ such that for any $\boldsymbol{x}, \boldsymbol{y} \in \mathbb{R}^E$ with $\|\boldsymbol{x} - \boldsymbol{y}\| \leq \frac{1}{\| \boldsymbol{L_1} \|_{\infty}}$ it holds $\forall j \in [E]$ that 
    \begin{align*}
        \left | \frac{\partial L(\boldsymbol{y})}{\partial x_j} - \frac{\partial L (\boldsymbol{x})}{\partial x_j} \right | \leq \left ( \frac{L_{0, j}}{\sqrt{E}} + L_{1, j} \left | \frac{\partial L (\boldsymbol{x})}{\partial x_j}  \right | \right ) \cdot ||\boldsymbol{y} - \boldsymbol{x}||_2.
    \end{align*}
\end{assumption}

This is a generalization of the scalar $(L_0, L_1)$-smoothness: 

\begin{definition}\label{D1}
    $L$ is called $(L_0, L_1)$-smooth if there exist scalars $L_0, L_1 \in \mathbb{R}$ such that for all $\boldsymbol{x} \in \mathbb{R}^E$ it holds that
    \begin{align*}
        ||\nabla^2 L(\boldsymbol{x})|| \leq L_0 + L_1 ||\nabla L(\boldsymbol{x})||.
    \end{align*}
\end{definition}

The coordinate-wise $(\boldsymbol{L_0}, \boldsymbol{L_1})$-smoothness implies that smoothness may vary for each coordinate of the input space. 
This particularly pertains to LLMs as it has been shown in \cite{robustness} that variance can be observed across mostly every transformer layer, such that each layer coordinate $j$ satisfies an own $(L_{0,j}, L_{1,j})$ pair. 
Thus, if the coefficients $L_{1,j}$ are non-zero, smoothness is potentially \textit{unbounded}. 
In contrast, if all $L_{1,j}$ are strictly zero, the original Lipschitz smoothness is recovered. 
In \cite{robustness}, the following Lemma has been established, relating the coordinate-wise smoothness to the loss divergence:

\begin{lemma}\label{L1}
    Let $L$ be $(\boldsymbol{L_0}, \boldsymbol{L_1})$-smooth coordinate-wisely. Then for any $\boldsymbol{x}, \boldsymbol{y} \in \mathbb{R}^E$ with $|| \boldsymbol{x} - \boldsymbol{y} ||_2 \leq \frac{1}{|| \boldsymbol{L_1} ||_{\infty}}$, we have
    \begin{align}
        L(&\boldsymbol{y}) \leq L(\boldsymbol{x}) + \langle \nabla L(\boldsymbol{x}), \boldsymbol{y} - \boldsymbol{x} \rangle \nonumber \\
        &+ \sum_{j=1}^E \frac{\left ( 
        \frac{L_{0, j}}{\sqrt{E}} + L_{1,j} \left | \frac{\partial L (\boldsymbol{x}) }{\partial x_j} \right | \right ) ||\boldsymbol{y} - \boldsymbol{x}||_2}{2} \ |y_j - x_j|.
    \end{align}
\end{lemma}

\subsubsection{Upper Bound on the LLM Loss Divergence} We utilize Lemma \ref{L1} to derive the loss divergence upper bound as follows.

\begin{lemma}\label{P0}
    For $\boldsymbol{x}, \boldsymbol{y} \in \mathbb{R}^E$, the loss divergence is bounded by 
    \begin{align}
        | L(\boldsymbol{y}) - L(\boldsymbol{x}&) | \leq \| \nabla L(\boldsymbol{x}) \|_2 \cdot \| \boldsymbol{y} - \boldsymbol{x} \|_2 \nonumber \\ 
        &\ + \left \| \boldsymbol{L_0} + \boldsymbol{L_1} \odot | \nabla L(\boldsymbol{x}) | \ \right \|_2 \cdot \| \boldsymbol{y} - \boldsymbol{x} \|_2^2. \label{eq:upper_bound_losses_without_tau_lemma_2}
    \end{align}

    \begin{proof}
        See Appendix \ref{app:A}
    \end{proof}

\end{lemma} 
  
In \eqref{eq:upper_bound_losses_without_tau_lemma_2}, the gradient loss $\| \nabla L(\boldsymbol{x}) \|_2$ might be unbounded, particularly when several coordinates need to be considered. 
However, common practice in deep learning suggests to bound gradients manually via gradient clipping \cite{zhang2019gradient} using a clipping threshold $\tau > 0$, thereby preventing exploding gradients, i.e.
\begin{align}
    \nabla L(\boldsymbol{x}) = 
        \begin{cases} 
            \nabla L(\boldsymbol{x}) & \text{, if } \| \nabla L(\boldsymbol{x}) \|_2 \leq \tau \\
            \frac{\tau}{\| \nabla L(\boldsymbol{x}) \|_2} \cdot \nabla L(\boldsymbol{x}) & \text{, otherwise }  
        \end{cases}
\end{align}

\begin{corollary}\label{R1}
    If gradient clipping is applied, the upper bound on the LLM loss divergence from Lemma \ref{P0} simplifies to
    \begin{align}
        \left | L(\boldsymbol{y}) - L(\boldsymbol{x}) \right | &\leq \tau \cdot \| \boldsymbol{y} - \boldsymbol{x} \|_2 \nonumber \\ &\ +\left \| \boldsymbol{L_0} + \tau \boldsymbol{L_1} \right \|_2 \cdot \| \boldsymbol{y} - \boldsymbol{x} \|_2^2. \label{eq:losses_upper_bound}
    \end{align}
\end{corollary}

Lemma \ref{P0} thus provides an upper bound on the divergence between loss functions for two distinct inputs $\boldsymbol{x}$ and $\boldsymbol{y}$. 
This allows us to quantify the impact of adversarial jamming by measuring the loss divergence between legitimate and corrupted inputs. 
Corollary \ref{R1} further refines this upper bound for practical applications by incorporating gradient clipping.

\subsubsection{Relating the Model Error to the Communication MSE}
Having established the necessary upper bounds on the loss divergence, we now apply those in the context of legitimate and jammed embeddings. 
To this end, we first show an equivalence between the embedding MSE and the communication MSE in Proposition \ref{P1}. 
We then use this equivalence in Proposition \ref{P2} to establish a direct relationship between the model error, expressed by the expected loss divergence, and jamming, quantifying the jammer's impact on the training performance.

\begin{proposition}\label{P1}
    Let $\boldsymbol{e}_q, \hat{\boldsymbol{e}}_q \in \mathbb{R}^E$ be the true and corrupted embeddings and let $s_{qnk}, \hat{s}_{qnk} \in \mathbb{C}$ be the corresponding true and corrupted transmit symbols. Then, it holds that
    \begin{align}
        \mathbb{E} \left [ \| \boldsymbol{e}_q - \hat{\boldsymbol{e}}_q \|_2^2 \right ] &= \sum_{(n,k) \in \mathcal{R}_q} \alpha_{qnk} \mu_{qnk} \\ &
        = \mathbb{E} \left [ \| \boldsymbol{s}_q - \hat{\boldsymbol{s}}_q \|_2^2 \right ],
    \end{align}
    where $\mu_{qnk}$ denotes the expected symbol error per resource allocation, i.e.
    \begin{align}
        \mu_{qnk} &= \mathbb{E} \left [ | s_{qnk} - \hat{s}_{qnk} |^2 \right ] \\
        &= | p_{qnk} \boldsymbol{v}_{qnk}^H \boldsymbol{H}_{qnk} \boldsymbol{w}_{qnk} - 1 |^2 + \boldsymbol{v}_{qnk}^H \boldsymbol{X}_{qnk} \boldsymbol{v}_{qnk} ,
    \end{align}
    with the expectation being taken over the joint distribution of $\{s_{qnk}\}_{(n,k) \in \mathcal{R}_q}$, $\boldsymbol{z}_{nk}$ being conditioned on $\boldsymbol{e}_q$, and with the interference-plus-noise covariance matrix $\boldsymbol{X}_{qnk}$, i.e.
    \begin{align}
        \boldsymbol{X}_{qnk} = \sum_{q' \neq q} \boldsymbol{H}_{q'nk} \boldsymbol{b}_{q'nk} \boldsymbol{b}_{q'nk}^H \boldsymbol{H}_{q'nk}^H + \boldsymbol{C}_{\boldsymbol{z}_{nk}}
        \label{eq:X}
    \end{align}
    with the shorthand $\boldsymbol{b}_{qnk} =  \alpha_{qnk} \sqrt{p_{qnk}} \boldsymbol{w}_{qnk}$ and composite noise covariance $\boldsymbol{C}_{\boldsymbol{z}_{nk}}$.

    \begin{proof}
        The proof follows directly from $s_{qnk}$ and $z_{qnk}$ being uncorrelated for all $q,n,k$, and from the assumption that we can recover $\{s_{qnk}\}_{(n,k) \in \mathcal{R}_q}$ from $\boldsymbol{e}_{qnk}$ and vice versa.
    \end{proof}

\end{proposition}

\begin{proposition}\label{P2}
    Let $L: \mathbb{R}^E \rightarrow \mathbb{R}$ satisfy Assumptions \ref{A1} and \ref{A2} with coordinate-wise smoothness parameters $\boldsymbol{L}_0, \boldsymbol{L}_1 \in \mathbb{R}^E$, and let $\boldsymbol{e}_q, \hat{\boldsymbol{e}}_q \in \mathbb{R}^E$ be the true and corrupted embeddings with $\|\boldsymbol{e}_q - \hat{\boldsymbol{e}}_q\| \leq \frac{1}{\| \boldsymbol{L_1} \|_{\infty}}$.
    Then, it holds that
    \begin{align}
        \mathbb{E} \left [ \left | L(\boldsymbol{e}_q) - L(\hat{\boldsymbol{e}}_q) \right | \right ] &\leq \| \nabla_{\boldsymbol{e}_q} L(\boldsymbol{e}_q) \|_2 \cdot \sqrt{\mathbb{E} \left [ \| \boldsymbol{s}_q - \hat{\boldsymbol{s}}_q \|_2^2 \right ]} \nonumber \\
        &\ + \| \boldsymbol{u}(\boldsymbol{e}_q) \|_2 \cdot \mathbb{E} \left [ \| \boldsymbol{s}_q - \hat{\boldsymbol{s}}_q \|_2^2 \right ], \label{eq:P2_upper_bound}
    \end{align}
    with the expectation being taken over the joint distribution of $\{s_{qnk}\}_{(n,k) \in \mathcal{R}_q}$, $\boldsymbol{z}_{nk}$ being conditioned on $\boldsymbol{e}_q$, and with
    \begin{align}
        \boldsymbol{u}(\boldsymbol{e}_q) = \boldsymbol{L_0} + \boldsymbol{L_1} \odot \nabla_{\boldsymbol{e}_q} L (\boldsymbol{e}_q). \label{eq:u_eq}
    \end{align}

    \begin{proof}
        See Appendix \ref{app:B}
    \end{proof}

\end{proposition}

\begin{corollary}\label{R2}
    In the case of gradient clipping with $\tau > 0$, the upper bound in \eqref{eq:P2_upper_bound} from Proposition \ref{P2} further simplifies to 
    \begin{align}
        \mathbb{E} [ | L(\boldsymbol{e}_q) - L(&\hat{\boldsymbol{e}}_q) | ] \leq \tau \cdot \sqrt{\mathbb{E} \left [ \| \boldsymbol{s}_q - \hat{\boldsymbol{s}}_q \|_2^2 \right ]} \nonumber \\
        &\ + \| \boldsymbol{L_0} + \tau \boldsymbol{L_1} \|_2 \cdot \mathbb{E} \left [ \| \boldsymbol{s}_q - \hat{\boldsymbol{s}}_q \|_2^2 \right ]. \label{eq:P2_upper_bound_clipped}
    \end{align}
\end{corollary}

\subsubsection{Practical Interpretation of Results}
Proposition \ref{P2} provides an upper bound on the model error, defined by the expected loss divergence between legitimate and corrupted embeddings, which is \emph{directly dependent} on the MSE of the wireless communication system. 
Therein, the proximity condition \( \| \boldsymbol{e}_q - \hat{\boldsymbol{e}}_q \|_2 \leq \frac{1}{\| \boldsymbol{L_1} \|_{\infty}} \) sets a practical constraint on the distance between legitimate and jammed embeddings, which needs to be small enough for the smoothness condition to hold. 
As outlined in \cite{robustness}, this ensures the stability of the gradient behavior, preventing numerical instabilities and unreliable approximations as gradient-dependent optimization algorithms might struggle to converge. 
By applying gradient clipping, we ensure that \( \| \nabla_{\boldsymbol{e}_q} L(\boldsymbol{e}_q) \|_2 \) is bounded by $\tau$, thereby stabilizing the training process as the assumptions underlying the optimization methods are not violated. 
In particular, Corollary \ref{R2} ensures that the model error does not explode and is only dependent on the $(\boldsymbol{L_0}, \boldsymbol{L_1})$-smoothness coefficients, the clipping threshold $\tau$, and the communication MSE $\mathbb{E}[ \| \boldsymbol{s}_q - \hat{\boldsymbol{s}}_q \|_2^2]$, independent of whether the proximity condition is fulfilled or not. 
In the context of \emph{jamming}, this allows for the consideration of arbitrary adversaries, including worst-case scenarios. 
Hence, even if the resulting jammed embeddings do not fulfill the proximity condition, for example due to excessive noise or sophisticated attack strategies that might flip the embedding label, our analysis remains applicable, aligning with best practices in deep learning. 
This \emph{new insight} establishes a formal relationship between the transformer-based LLM architecture, its semantic embeddings, and the wireless medium, thereby emphasizing the importance of the wireless communication system and its \emph{resilience} to adversarial jamming in the quality of distributed training. 
To the best of our knowledge, this is the \emph{first} formal characterization of such a relationship for practical DCML with LLMs over wireless networks. 
In particular, we consider a generalized smoothness assumption on the loss function, which is often omitted in previous works but required for a proper analysis. 
This assumption explains why techniques such as gradient clipping work and might be necessary during training, and how corrupted model inputs affect the loss divergence, a critical measure for the ML training performance. 
Consequently, as the model error directly depends on the communication MSE, a \emph{wireless} approach to resilience in SFL is not only justified but required, thus instructing us to maximize the SINR.
\subsection{Minimum System Rate for Reliable SFL with LLMs.}
\label{sec:minimum_system_rate}
To characterize the minimum system rate under which the communication link can support SFL reliably, we need to identify outage conditions caused by jamming. 
To this end, we provide three remarks. 
In Remark 1, we first derive a lower bound for the per resource allocation symbol error. We use this lower bound in Remark 2 to establish a general lower bound on the outage rate, which is dependent on the $\boldsymbol{L}_1$ constant from Lemma \ref{L1}. 
In Remark 3, we derive the outage rate for the particular case where Proposition \ref{P2} is fulfilled with equality.
\begin{remark}\label{O1}
    Let the resource set for each user $q$ be defined as $\mathcal{R}_q = \{ (n,k) | \alpha_{qnk} \neq 0 \ \forall (n,k) \in \mathcal{R} \}$ with $|\mathcal{R}_q| = r_q$. For the minimum MSE (MMSE) receive filter $v_{qnk}$ with
    \begin{align}
        v_{qnk} = (\boldsymbol{X}_{qnk} + \boldsymbol{H}_{qnk} \boldsymbol{b}_{qnk} \boldsymbol{b}_{qnk}^H \boldsymbol{H}_{qnk}^H )^{-1} \boldsymbol{H}_{qnk} \boldsymbol{b}_{qnk}, \label{eq:mmse_rx}
    \end{align}
    we have the following well-known equality for the per resource allocation symbol error $\mu_{qnk}$:
    \begin{align}
        \mu_{qnk} = \exp \{-I (s_{qnk}, \hat{s}_{qnk})\} = \exp \{- R_{qnk}\}. \label{eq:mse_inequality}
    \end{align}

    Using \eqref{eq:mse_inequality}, we further have $\mu_q = \sum_{(n,k) \in \mathcal{R}_q} \exp \{-R_{qnk}\}$, and then, we can apply Jensen's inequality to obtain
    \begin{align}
        \mu_q \geq r_q \cdot \exp \left \{-r_q^{-1} \sum_{( n,k) \in \mathcal{R}_q} R_{qnk} \right \} = r_q \cdot \exp \{-r_q^{-1} R_q\} \label{eq:o1}.
    \end{align}
\end{remark}

\begin{remark}\label{O2}
    In Proposition \ref{P2}, we require $\|\boldsymbol{e}_q - \hat{\boldsymbol{e}}_q\| \leq \frac{1}{\| \boldsymbol{L_1} \|_{\infty}} \ \forall \boldsymbol{e}_q, \hat{\boldsymbol{e}}_q \in \mathbb{R}^E$, which further implies that 
    \begin{align}
        \mathbb{E} [ \| \boldsymbol{e}_q - \hat{\boldsymbol{e}}_q \|_2^2] = \mu_q \leq \mathbb{E}\left [ \| \boldsymbol{L_1} \|_{\infty}^{-2} \right ] = \| \boldsymbol{L_1} \|_{\infty}^{-2}.
    \end{align}

    Using \eqref{eq:o1}, we may conclude that
    \begin{align}
        r_q \cdot \exp \{-r_q^{-1} R_q\} \leq \mu_q \leq \| \boldsymbol{L_1} \|_{\infty}^{-2}.
    \end{align}

    From here, it is easy to derive a lower bound for $R_q$ as 
    \begin{align}
        R_q \geq r_q \cdot \log \left ( \| \boldsymbol{L_1} \|_{\infty}^{2} r_q \right ) \stackrel{\mathclap{def}}{=} R_{\textrm{out}, 1},
    \end{align}
    where $R_{\textrm{out}, 1}$ represents the minimum rate required for Proposition \ref{P2} to hold in its expectation.
\end{remark}

\begin{remark}\label{O3}
    We are interested in the rate $R_{\textrm{out}, 2}$, for which \eqref{eq:P2_upper_bound} in Proposition \ref{P2} is fulfilled with equality. 
    To this end, we simplify its notation for ease of analysis as follows
    \begin{align}
        \mathbb{E} \left [ \left | L(\boldsymbol{e}_q) - L(\hat{\boldsymbol{e}}_q) \right | \right ] = \epsilon_q \leq \gamma_q \sqrt{\mu_q} + \delta_q \mu_q, 
    \end{align}
    where $\epsilon_q, \gamma_q, \delta_q$ are substitutes for the corresponding expressions in \eqref{eq:P2_upper_bound}, and $\mu_q$ as in \eqref{eq:o1} from Remark \ref{O1}. 
    Now, we require $\epsilon_q = \gamma_q \sqrt{\mu_q} + \delta_q \mu_q$ with equality. 
    By setting $y = \sqrt{\mu_q} = \sqrt{r_q \exp\{\-R_q/r_q\}}$, we can equivalently state that
    \begin{align}
        \delta_q y^2 + \gamma_q y - \epsilon_q = 0,
    \end{align}
    which has only one non-negative solution:
    \begin{align}
        y = \frac{\gamma_q}{2 \delta_q} \left ( \sqrt{1 + \frac{1 + 4 \gamma_q \epsilon_q}{\delta_q^2}} -1 \right ).
    \end{align}
    Thus, we obtain $R_{\textrm{out}, 2}$ by solving $y^2 = r_q \cdot \exp\{\frac{R_q}{r_q}\}$ for $R_q$:
    \begin{align}
        R_{\textrm{out}, 2} = r_q \cdot \log (r_q / y^2).
    \end{align}

    In the context of SFL, $R_{\textrm{out}, 2}$ represents the rate for user $q$ at which we can train reliably up to an error $\epsilon_q$.   
\end{remark}

From Remarks \ref{O2} and \ref{O3}, we can conclude that both the communication system and SFL perform reliably for system rates $R > R_{\textrm{out}} = \min\{R_{\textrm{out}, 1}, R_{\textrm{out}, 2}\}$. 
Note that the minimum rate $R_{\textrm{out}}$ depends on either $\boldsymbol{L}_1$ from to the proximity condition or on the particular realization of the adversarial jammer captured by the MSE. 
Thus, particularly strong or \emph{worst-case} jammers may be able to considerably decrease the system rate below $R_{\textrm{out}}$, such that neither automatic repeat requests (ARQs) nor other upper-layer resilience mechanisms can be applied. 
As $R_{\textrm{out}}$ scales with the number of users and available bandwidth, robust anti-jamming and resource allocation become essential for sustaining performance in larger SFL deployments.
To preserve SFL training under diverse and unpredictable disruptions, we specifically need to demand more than mere robustness, which only accounts for moderate or known distribution shifts, and instead require resilient defenses.    
This advocates the need for \emph{proactive}, resilient-by-design physical layer solutions to anti-jamming to ensure resilient and reliable wireless SFL, even under worst-case conditions where reactive network defenses cannot suffice. 
Moreover, the proximity condition involving $\boldsymbol{L}_1$ further indicates that the system can still maintain resilience if the jammer introduces only moderate noise, thereby not violating smoothness as long as the embeddings are not too divergent. 
This insight further opens up the door for \emph{adversarial training} aspects under which resilience can even be increased due to controlled noise exposure \cite{jain2023neftune}, which we verify in Section IV.

\section{R-SFLLM Anti-Jamming Framework}
\label{sec:r_sfllm_framework}
In this section, we develop the R-SFLLM anti-jamming component. 
To this end, we first define the anti-jamming optimization problem and provide insights into the role of sensing-assisted DoA information. 
Then, we solve the optimization problem using an iterative water-filling solution. 
Finally, we provide an analytical expression for the worst-case jamming strategy as a benchmark for our SFL resilience framework.

\subsection{Anti-Jamming Strategy and Optimization Problem.}
\label{sec:anti_jamming_strategy}
As a result of Proposition \ref{P2}, jammed embeddings $\hat{\boldsymbol{e}}_q$ lead to a deviation from the ground truth in the deterministic loss function $L$. 
Thus, the anti-jamming objective can be generally formulated as the minimization of the expected loss divergence, i.e. $\min \ \mathbb{E} \left [ \left | L(\boldsymbol{e}_q) - L(\hat{\boldsymbol{e}}_q) \right | \right ]$. 
Using Corollary \ref{R2}, we can instead minimize $J(\boldsymbol{s}_q, \hat{\boldsymbol{s}}_q) = \sqrt{\mathbb{E} \left [ \| \boldsymbol{s}_q - \hat{\boldsymbol{s}}_q \|_2^2 \right ]} + \mathbb{E} \left [ \| \boldsymbol{s}_q - \hat{\boldsymbol{s}}_q \|_2^2 \right ]$, which is only dependent on the MSE and where we imply using gradient clipping during training. 
This problem can be equivalently interpreted as the maximization of the SINR for each user $q \in \mathcal{Q}$, respectively, or more generally, as the maximization of the sum rate. To this end, we derive an expression for the achievable sum rate as follows:   
\begin{align}
    R &= \sum_{(n,k) \in \mathcal{R}_q \forall q} I(\boldsymbol{y}_{nk}; \{s_{qnk}\}_{q \in \mathcal{Q}}) \\
    &= \sum_{(n,k) \in \mathcal{R}_q \forall q} \log \left ( 1 + \sum_{q \in \mathcal{Q}} \alpha_{qnk} p_{qnk} \gamma_{qnk} (\boldsymbol{C}_{\boldsymbol{z}_{nk}}) \right ). \label{eq:transmitter_side_sum_rate}
\end{align}

In this setup, $\gamma_{qnk} (\boldsymbol{C}_{\boldsymbol{z}_{nk}})$ represents the SINR of user $q$ for the composite noise covariance matrix $\boldsymbol{C}_{\boldsymbol{z}_{nk}}$ and allocated resource elements $(n, k) \in \mathcal{R}_q$, i.e.
\begin{align}
    \gamma_{qnk}(\boldsymbol{C}_{\boldsymbol{z}_{nk}}) = \boldsymbol{w}_{qnk}^H \boldsymbol{H}_{qnk}^H \boldsymbol{C}_{\boldsymbol{z}_{nk}}^{-1} \boldsymbol{H}_{qnk} \boldsymbol{w}_{qnk}.
    \label{eq:gamma}
\end{align}

To incorporate anti-jamming in SFL \emph{by-design}, we need to jointly optimize over beamforming, user scheduling, and resource allocation constraints, thus introducing resilience \emph{proactively} at the bit level. 
To this end, we pose the following optimization problem applied for $q \in \mathcal{Q}, n \in \mathcal{N}$, and $k \in \mathcal{K}$: 
\begin{alignat}{2}
    \max_{
    \substack{\alpha_{qnk}, p_{qnk}, \boldsymbol{w}_{qnk}, \boldsymbol{v}_{qnk} \\
    (n, k) \in \mathcal{R}_{q}, q \in \mathcal{Q}}
    } \quad &R \quad \text{s.t.} \quad \label{eq:P1} \\
    \alpha_{qnk} \quad &\in \quad \{0,1\}, \tag{36a} \label{eq:C1} \\
    \alpha_{qnk} p_{qnk} \quad &\geq \quad 0,\tag{36b} \label{eq:C2} \\
    \sum_{(n,k)\in\mathcal{R}_{q}}\alpha_{qnk} \quad &\leq \quad B_{q},\tag{36c} \label{eq:C3}\\ 
    \sum_{(n,k)\in\mathcal{R}_{q}}\alpha_{qnk}p_{qnk} \quad &\leq \quad P_{q},\tag{36d} \label{eq:C4}\\
    \|\boldsymbol{w}_{qnk}\|_2^2 \quad &\leq \quad 1. \tag{36e} \label{eq:C5}
\end{alignat}

In this problem, constraints \eqref{eq:C1} to \eqref{eq:C3} incorporate the binary user scheduling, power allocation and resource block limitation, with $B_q = |\mathcal{R}_q|$ being the maximum number of resource allocation blocks for each user $q$. 
Constraint \eqref{eq:C4} captures the power budget and constraint \eqref{eq:C5} ensures the unit normalization of the beamforming vector. 
However, due to the binary user scheduling variable $\alpha_{qnk}$, the optimization is a non-linear, non-convex mixed-integer problem and thus NP-hard in general. 
In addition to NP-hardness, the sum rate further depends on the composite noise covariance matrix $\boldsymbol{C}_{\boldsymbol{z}_{nk}}$. 
Since the latter characterizes the adversarial jamming strategy, which is not known to the legitimate SFL parties, a surrogate expression $\widetilde{\boldsymbol{C}}_{\boldsymbol{z}_{nk}}$ needs to be found that effectively approximates the true covariance, i.e. $\widetilde{\boldsymbol{C}}_{\boldsymbol{z}_{nk}} \approx \boldsymbol{C}_{\boldsymbol{z}_{nk}}$.
\subsection{Role of Sensing-Assisted Jamming DoA Information.}
\label{sec:role_of_doa}
We have shown in \cite{andrei_jcs_symp23} that such a surrogate covariance can be approximated using the jamming signal DoAs as follows:
\begin{align}
    \widetilde{\boldsymbol{C}}_{\boldsymbol{z}_{nk}} = \eta \boldsymbol{A}(\boldsymbol{\theta}_G) \boldsymbol{A}(\boldsymbol{\theta}_G)^H + \sigma^2 \boldsymbol{I}_{N_R} \succeq \boldsymbol{C}_{\boldsymbol{z}_{nk}} \label{eq:surrogate}
\end{align}
with the array manifold evaluated at the known DoAs, i.e.
\begin{align}
    \boldsymbol{A}(\boldsymbol{\theta}_G) = \left [ \boldsymbol{a}_{N_R}(\boldsymbol{\theta}_{G, 1}) \ \hdots \ \boldsymbol{a}_{N_R}(\boldsymbol{\theta}_{G, L_G}) \right ].
\end{align}

This was motivated by showing that the true SINR $\gamma$ can be lower bounded by an approximate expression $\Tilde{\gamma}$, which is dependent on a scaling parameter $\eta$ and the DoAs as follows:
\begin{align}
    \gamma(\boldsymbol{w}, \boldsymbol{v}) \geq \frac{\boldsymbol{v}^H \boldsymbol{H} \boldsymbol{w} \boldsymbol{w}^H \boldsymbol{H}^H \boldsymbol{v}}{\boldsymbol{v}^H (\eta \boldsymbol{A}_{Rx}(\boldsymbol{\theta}_G) \boldsymbol{A}_{Rx} (\boldsymbol{\theta}_G)^H + \sigma^2 \boldsymbol{I}) \boldsymbol{v}} \stackrel{\mathclap{def}}{=} \Tilde{\gamma}(\boldsymbol{w}, \boldsymbol{v}).
\end{align}

By inserting \eqref{eq:surrogate} into \eqref{eq:transmitter_side_sum_rate}, we hence obtain a lower bound on $R$. 
Note that in general $\eta$ is unknown since it depends on the unknown jamming setup. 
Thus, we consider it as a conscious \emph{hyperparameter}, which controls the resilience level of our system. In \cite{andrei_jcs_symp23}, we showed that by choosing $\eta$ to be much larger than the noise level $\sigma^2$, i.e. $\eta \gg \sigma^2$, we coincide with the case where the jammer setup is known, that is where $\gamma \approx \Tilde{\gamma}$. 
In this case, the SINR can be maximized by maximizing the lower bound $\Tilde{\gamma}$ instead. 
Thus, $\widetilde{\boldsymbol{C}}_{\boldsymbol{z}_{nk}}$ constitutes a conservative approximation of $\boldsymbol{C}_{\boldsymbol{z}_{nk}}$, ensuring it does not underestimate the impact of noise and adversarial jamming, indicated by the Löwner order $\succeq$ in \eqref{eq:surrogate}. 
We carefully validated this conjecture in \cite{andrei2024resilientbydesign} for $\eta = 10$ in several jamming scenarios for a range of power budgets $P_J < \infty$. 
Consequently, the availability of the DoAs alleviates the need to know the exact jamming statistics.
In future 6G networks, having access to jamming DoAs is a \emph{realistic} assumption, as native wireless sensing services can provide this information, for example, as part of ISAC and RIS protocols. We refer the interested reader to a comprehensive overview of such protocols in \cite{zhu2024enabling} and \cite{gonzalez2024integrated}.
\begin{algorithm}[t]
    \footnotesize
    \caption{Joint Iterative Scheduling, Beamforming, and Power Allocation}
    \label{alg:IWF_0}

    \KwIn{Legitimate channel $\boldsymbol{H}_{qnk}$, noise covariance $\boldsymbol{C}_{\boldsymbol{z}_{nk}}$, power constraint $P_q$, maximum number of resource allocation blocks $B_q$}
    \KwOut{Wireless system design variables $\alpha_{qnk}$, $p_{qnk}$, $\boldsymbol{w}_{qnk}$}

    \vspace{1mm} \hrule \vspace{1mm}

    Initialize the interference-plus-noise covariance matrix to $\boldsymbol{X}_{qnk} = \boldsymbol{C}_{\boldsymbol{z}_{nk}}$ \\
    Initialize the design variables $\alpha_{qnk}^0$, $p_{qnk}^0$, $\boldsymbol{w}_{qnk}^0$ using the single user update procedure in steps (6)-(11) \\

    \While{not converged}{
        \For{$q = 1$ to $Q$}{
            Update $\boldsymbol{X}_{qnk}$ using Equation \eqref{eq:X} \\
            Compute the \textit{maximum eigenvalues} $\{\lambda_{qnk}\}_{(n,k) \in \mathcal{R}_q}$ of $\{\boldsymbol{H}_{qnk}^H \boldsymbol{X}_{qnk}^{-1} \boldsymbol{H}_{qnk}\}_{(n,k) \in \mathcal{R}_q}$ \\
            Compute the corresponding eigenvectors $\{\boldsymbol{u}_{qnk}\}_{(n,k) \in \mathcal{R}_q}$ \\
            Determine the indices $\mathcal{I}_q$ of the largest $B_q$ eigenvalues \\
            Set the user scheduling to $\alpha_{qnk} = 1$ for all $(n,k) \in \mathcal{I}_q$ and 0 otherwise \\
            Compute the power allocation $p_{qnk} = (\mu - \lambda_{qnk}^{-1})^{+}$ with $\mu$ chosen such that $\sum_{\mathcal{I}_q} p_{qnk} \leq P_q$ \\
            Set the beamforming vector to $\boldsymbol{w}_{qnk} = \boldsymbol{u}_{qnk}$ for $(n,k) \in \mathcal{I}_q$
        }
    }
\end{algorithm}

\subsection{Iterative Water-Filling Solution.}
\label{sec:water_filling_solution}
We have further shown in \cite{andrei2024resilientbydesign} that the NP-hard problem in \eqref{eq:P1} can be iteratively solved using a water-filling approach, described in \refalg{alg:IWF_0}. 
The proposed method adapts the original water-filling for MAC and MIMO channels \cite{mimo_iterative_waterfilling} to incorporate user scheduling. 
To this end, each user $q \in \mathcal{Q}$ determines its update on the matrix $\boldsymbol{X}_{qnk}$ and computes an optimal set of the wireless system design parameters $\alpha_{qnk}$, $p_{qnk}$, $\boldsymbol{w}_{qnk}$, as outlined in steps (6)-(11), where for each user the following optimization problem is solved:
\begin{align}
    &\max_{\substack{\alpha_{qnk}, p_{qnk}, \\ \boldsymbol{w}_{qnk}}} \sum_{(n,k) \in \mathcal{R}_q} \underbrace{ \alpha_{qnk} \cdot \log \left ( 1 + p_{qnk} \gamma_{qnk} (\boldsymbol{X}_{qnk}) \right ) }_{R_q}.
\end{align}

\refalg{alg:IWF_0} effectively circumvents the need to deal with the NP-hardness of the problem as the overall sum rate is indirectly maximized by maximizing the sum rate $R_q$ of the strongest users individually.
In each iteration, we use the power iteration method for computing eigenvalues and eigenvectors, and perform water-filling for power allocation. 
Both of these methods are known to exhibit rapid convergence. 
Consequently, \refalg{alg:IWF_0} inherits these convergence properties, which we verified in extensive experiments, where our method on average takes less than five iterations to converge. 
With $n_{\text{iter}}$ representing the number of iterations required for convergence, the overall complexity of the algorithm is
\begin{align}
    \mathcal{O}(n_{\text{iter}} Q N K N_R N_T^2).
\end{align} 

Thus, the complexity of our anti-jamming solution increases polynomially with the number of users $Q$, hence remaining scalable for most realistic SFL deployments. We will illustrate this in our experiments in Section \ref{sec:experiments}.
\subsection{Worst-Case Jamming Strategy.}
\label{sec:wc_strategy}
In Proposition \ref{P2}, we have seen that jamming embeddings affects the training performance. 
However, the impact still remains to be investigated for worst-case conditions, and in particular, how this affects the global performance in SFL after aggregating such corrupted models. 
To this end, we need to derive the worst-case jamming strategy, which we use to benchmark R-SFLLM. In contrast to anti-jamming, we need to find the covariance matrix $\boldsymbol{C}_{\boldsymbol{u}_{nk}}$, which minimizes $R$ instead, and pose the following adversarial objective: 
\begin{alignat}{2}
    \min_{
    \substack{\boldsymbol{C}_{\boldsymbol{u}_{nk}}}
    } \quad &R \quad \text{s.t.} \quad \forall (n,k) \in \mathcal{R}_q \label{eq:P3} \\
    \boldsymbol{C}_{\boldsymbol{u}_{nk}} &= \boldsymbol{C}_{\boldsymbol{u}_{nk}}^H, \tag{42a} \label{eq:J1} \\
    \boldsymbol{C}_{\boldsymbol{u}_{nk}} &\succeq \boldsymbol{0}, \tag{42b} \label{eq:J2} \\
    \sum_{(n,k)\in\mathcal{R}_q} \tr \left ( \boldsymbol{C}_{\boldsymbol{u}_{nk}} \right ) &\leq P_J. \tag{42c} \label{eq:J3}
\end{alignat}

This is a convex semidefinite program due to the convex objective function and constraints. 
Thus, a global optimum can be found in polynomial time using interior-point or first-order methods \cite{boyd_cvx}. 
However, neither one of these approaches scale efficiently for high-dimensions with a complexity of $\mathcal{O}(N_J^4 N^2 K^2)$, or higher \cite{wolkowicz2012handbook}. 
Thus, we require a compute-efficient alternative. To this end, we proposed a two-step approximation procedure in \cite{andrei2024resilientbydesign} described in \refalg{alg:WCJ}, which consists of a prior user selection stage and a subsequent compute-efficient convex optimization. 
In \cite{andrei2024resilientbydesign}, we have further shown that this worst-case jammer nullifies the sum rate, independent of the number of antennas or DoAs. 
Thus, we adopt this adversary as a benchmark in the subsequent experiments and investigate its impact on global model performance in SFL with LLMs to answer the question of whether worst-case jamming translates into worst-case SFL training performance.

\begin{algorithm}[t]
    \footnotesize
    \caption{Approximate Worst-Case Jamming Strategy}
    \label{alg:WCJ}

    \KwIn{Legitimate channel $\boldsymbol{H}_{qnk}$, jamming channel $\boldsymbol{G}_{nk}$, jamming power budget $P_J$}
    \KwOut{Worst-case jamming covariance matrix $\boldsymbol{C}_{\boldsymbol{u}_{nk}}$}

    \vspace{1mm} \hrule \vspace{1mm}

    \For{each user $q \in \mathcal{Q}$}{
        Compute the alignment matrix $\boldsymbol{R}_{qnk} = \boldsymbol{G}_{nk}^{\dagger} \boldsymbol{H}_{qnk} \boldsymbol{H}_{qnk}^H \boldsymbol{G}_{nk}^{\dagger, H}$ \\
    }
    
    Determine the strongest user $q^* = \arg \max_{q \in \mathcal{Q}} \lambda_{\max}(\boldsymbol{R}_{qnk})$ \\
    
    \For{the strongest user alignment $\boldsymbol{R}_{q^*nk}$}{
        Compute the eigenvector matrix $\boldsymbol{U}_{q^*nk}$ and eigenvalues $\{\lambda_{q^*nk,d}\}_{d=1}^{N_J}$ \\
    }
    
    Compute the jamming power allocation weighting $h_{nk} = p_{q*nk} \sum_{d=1}^{N_J} \lambda_{q*nk,d}$ \\
    
    Compute the jamming power scaling factors $g_{nk} = \left( P_J \sqrt{h_{nk}} \right) / \left( \sum_{nk} \sqrt{h_{nk}} \right)$ \\
    
    Compute the jamming power allocation matrix $\boldsymbol{\Lambda}_{\boldsymbol{u}_{nk}} = \text{diag}\left\{ \dfrac{g_{nk} \lambda_{q^*nk, d}}{\sum_{d=1}^{N_{J}} \lambda_{q^*nk, d}}  \right\}_{d=1}^{N_{J}}$ \\
    
    Compute the jamming covariance matrix $\boldsymbol{C}_{\boldsymbol{u}_{nk}} = \boldsymbol{U}_{q^*nk} \boldsymbol{\Lambda}_{\boldsymbol{u}_{nk}} \boldsymbol{U}_{q^*nk}^H$
    
\end{algorithm}

\section{Experiments, Simulation Results, and Analysis}
\label{sec:experiments}

In this section, we discuss our simulation results for applying R-SFLLM to SFL training with language models. 
We provide insights into the sensitivity of SFL to poisoned model aggregations, the impact of the worst-case jammer compared to barrage jamming, the effectiveness of our sensing-assisted anti-jamming strategy, and the influence of different model architectures.
We first present detailed results for the NLP domain when using LLMs in \refsec{sec:simulation_results_LLM}, and then extend our analysis to VLMs in \refsec{sec:simulation_results_VLM}, covering two modalities across 13 diverse datasets.
Finally, we present ablation results for varying the number of users $Q$ and the number of jamming antennas $N_J$ in R-SFLLM, demonstrating that our proposed framework is both scalable and effective.

\subsection{Experimental Setup.}
\label{sec:experimental_setup}

\begin{figure}[htbp]
    \centering
    \includegraphics[width=0.75\linewidth]{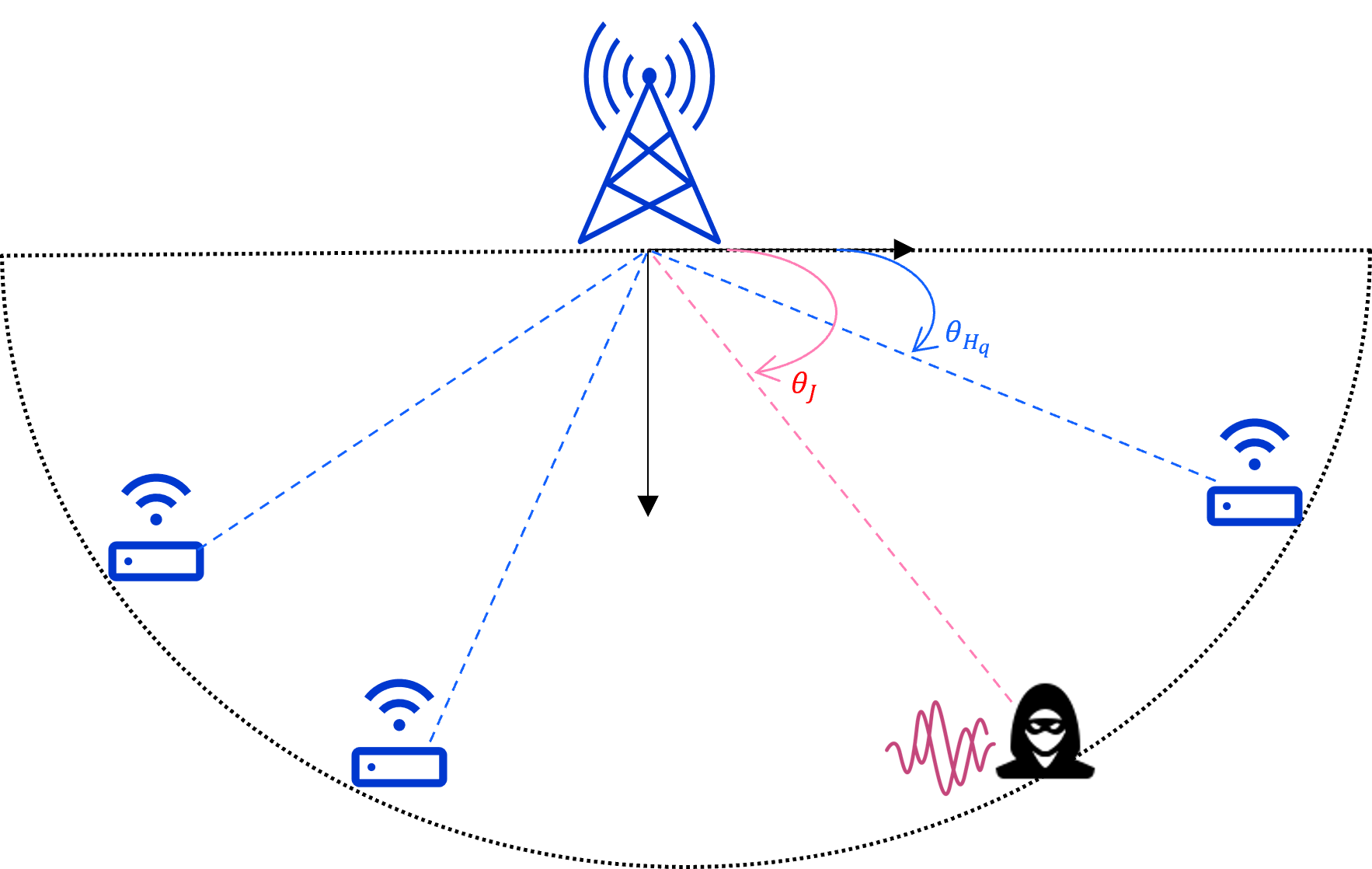}
    \caption{SFL setup with $Q = 3$ legitimate parties, one adversarial jammer, and corresponding user and jamming DoAs, denoted by $\theta_{H_q}$ and $\theta_J$, respectively.}
    \label{fig:jamming_setup}
\end{figure}

We refer to the R-SFLLM setup in \refig{fig:jamming_setup}, where $Q=3$ legitimate clients participate in \emph{fine-tuning} LLMs and VLMs for various NLP and CV tasks and datasets.

\subsubsection{NLP Experiments} We fine-tune BERT \cite{devlin2018bert} and RoBERTa \cite{liu2019roberta} base models for two distinct NLP tasks: Sequence classification (SC) and named entity recognition (NER).
SC assigns a category to a sequence of words or tokens, while NER identifies named entities, such as persons or organizations within a text. 
For SC, the binary classification datasets SST2 (67k samples) \cite{sst2} and QNLI (105k samples) \cite{wang2019glue}, as well as the ternary dataset MNLI (393k samples) \cite{mnli} are considered, while for NER CONLL2003 (14k samples) \cite{conll2003} and WNUT17 (3k samples)  \cite{wnut17} are used.
These datasets contain various text corpora ranging from news articles to movie reviews, thus spanning a wide range of information. 
For example, when fine-tuning on SST2, we perform sentiment analysis where the LLM is used to determine whether a movie review is positive or not. 
Each dataset is divided equally among the clients, ensuring a unique and private portion of the data. 
The varying dataset sizes allow us to further analyze the importance of the number of corrupted data points in SFL.

\subsubsection{CV Experiments} We follow the experimental setup in \cite{rmtllmf} and fine-tune OpenAI's CLIP ViT-B/16 VLM \cite{CLIP} with batch size 128 on eight different image classification datasets: 
MNIST (60k samples) \cite{mnist}, Cars (8k samples) \cite{cars}, DTD (4k samples) \cite{dtd}, EuroSAT (21k samples) \cite{eurosat}, GTSRB (27k samples) \cite{gtsrb}, RESISC45 (19k samples) \cite{resisc}, SUN397 (20k samples) \cite{sun}, and SVHN (73k samples) \cite{svhn}.
This variety in image data distributions allows us to investigate the impact of corruptions on additional vision modalities during training.  

For both NLP and CV, the pre-trained LLMs are fine-tuned for $N_{\text{epochs}} = 10$ epochs and $N_{\text{rounds}} = 10$ global SFL rounds.

\subsubsection{Adversarial Jamming} In our SFL setup, all participating parties employ uniform linear antenna arrays with $N_{T_q} = 8$ and $N_R = 16$ legitimate transmit and receive antennas. 
The adversary is assumed to be in the jammer-dominant regime and employs the worst-case jamming strategy from \refsec{sec:wc_strategy} to maximally corrupt the embeddings during each uplink transmission in the MIMO-OFDM MAC. 
The corresponding user and jamming DoAs are determined as
\begin{align}
    \theta_{H_{q,l}} = \theta_{H_{q}} + \phi_{H_{q,l}} \quad \text{and} \quad
    \theta_{G,l} = \theta_{J} + \phi_{G,l} 
\end{align}
where the central angles $\theta_{H_q}$ and $\theta_{J}$ are set to $0^\circ$ and $20^\circ$, respectively. 
The disturbance in form of the angle spread $\phi_{.,l}$ for each antenna $l$ is drawn uniformly via $\mathcal{U}(.)$ according to
\begin{align}
    \phi_{H_{q,l}} \sim \mathcal{U}[- 10^\circ, 10^\circ] \quad \text{and} \quad
    \phi_{G,l} \sim \mathcal{U}[- 5^\circ, 5^\circ].
\end{align}

The considered communication protocol employs 5G New Radio (NR) slots with $K=14$ symbols per slot and $N=64$ subcarriers. 
The maximum number of resource allocation blocks for each user $q$ is given by $B_q = \left \lfloor \frac{NK}{Q} \right \rfloor = 298$.
We choose the scale parameter of the sensing-assisted R-SFLLM anti-jamming to be $\eta = 10$, which is much larger than the background noise of $\sigma^2 = -3 \text{ dBm}$. 
This setup follows our work in \cite{andrei2024resilientbydesign}, however, different configurations can be applied without loss of generality.
We further resample the jamming statistics after each uplink transmission, i.e. after each training batch, to simulate movement and jamming variance. 

\quad 

The following four scenarios are studied as benchmarks:
\begin{enumerate}[leftmargin=*]
    \item \textit{SFL Baseline}: SFL performance without wireless model.
    
    \item \textit{Gaussian}: No adversarial jamming, only AWGN.
    
    \item \textit{No Protection}: Worst-case jamming without R-SFLLM.
    
    \item \textit{Protection}: Worst-case jamming with R-SFLLM.
\end{enumerate}

Table \ref{tab:sfl_config_params} summarizes the experiment configuration and Table \ref{tab:sfl_results} shows the fine-tuning results for each NLP and CV experiment scenario.
For our NLP experiments, we further include detailed performance plots for the aggregated global SFL model after each global round, i.e., after each $N_{\text{epochs}}$, for every dataset, base model, and scenario in \refig{fig:jointplot}.
For our CV experiments, we include a similar performance analysis in \refig{fig:vlm_plot}, averaged for all considered datasets. 
We provide relevant code and datasets as part of our GitHub repository\textsuperscript{\ref{repo}}. 

\begin{table}[t]
\centering
\resizebox{0.5\textwidth}{!}{%
\begin{tabular}{|l|l|l|}
\hline
\multicolumn{3}{|c|}{\textbf{Wireless Configuration Parameters}} \\ \hline
\( Q \) & Number of legitimate users & 3 \\ \hline
\( N_{T_q} \) & Number of legitimate transmit antennas per user & 8 \\ \hline
\( N_R \) & Number of legitimate receive antennas at the base station & 16 \\ \hline
\( N_J \) & Number of jammer transmit antennas & 64 \\ \hline
\( P_J \) & Jamming power & 30 dBm \\ \hline
\( P_q \) & Power of legitimate user \( q \) & 5 dBm \\ \hline
\( \theta_{H_{q,l}} \) & DoA for legitimate users & \( \theta_{H_{q}} + \phi_l \) \\ \hline
\( \theta_{G,l} \) & DoA for adversarial jammer & \( \theta_{J} + \phi_l \) \\ \hline
\( \theta_{H_q} \) & Central DoA for legitimate user & $0^\circ$ \\ \hline
\( \theta_J \) & Central DoA for adversarial jammer & $20^\circ$ \\ \hline
\( \phi_{H_{q,l}} \) & Angle spread for legitimate users & \( \pm 10^\circ \) \\ \hline
\( \phi_{G,l} \) & Angle spread for adversarial jammer &  \( \pm 5^\circ \) \\ \hline
\( K \) & Symbols per LTE-like slot & 14 \\ \hline
\( N \) & Subcarriers & 64 \\ \hline
\( B_q \) & Maximum number of resource blocks for user \( q \) & \( \lfloor \frac{NK}{Q} \rfloor = 298 \) \\ \hline
\( \eta \) & Scale parameter of anti-jamming framework & 10 \\ \hline
\( \sigma^2 \) & Background noise & -3 dBm \\ \hline
\( \text{Path loss} \) & Path loss & 10 dB \\ \hline
\( \text{Number of paths} \) & Number of propagation paths & 128 \\ \hline
\( f_c \) & Carrier frequency & 2.4 GHz \\ \hline

\multicolumn{3}{|c|}{\textbf{SFL Training Parameters}} \\ \hline
\( N_{\text{epochs}}, N_\text{rounds} \) & Number of training epochs \& global SFL rounds & 10 \\ \hline
$b_{\text{SST2, MNLI}}$ & Batch Size for the SST2 \& MNLI dataset & 64 \\ \hline
$b_{\text{QNLI}}$ & Batch Size for the QNLI dataset & 32 \\ \hline
$b_{\text{CONLL2003}}$ & Batch Size for the CONLL2003 dataset & 16 \\ \hline
$b_{\text{WNUT17}}$ & Batch Size for the WNUT\_17 dataset & 4 \\ \hline
$b_{\text{CV\_DATASETS}}$ & Batch Size for the CV experiment datasets & 128 \\ \hline
$\alpha$ & Learning rate & 1e-5 \\ \hline
$\epsilon$ & ADAM numerical stability constant & 1e-6 \\ \hline
$P_{\text{WARMUP}}$ & Warmup period in percent of iterations & 10\% \\ \hline
\end{tabular}%
}
\caption{Summary of SFL and wireless configuration parameters.}
\label{tab:sfl_config_params}
\vspace{-0.5cm}
\end{table}
\subsection{R-SFLLM Simulation Results for NLP Models.}
\label{sec:simulation_results_LLM}

For our NLP experiments, we separately evaluate SC and NER tasks, comparing BERT and RoBERTa model architectures and examining how dataset size affects performance.
In addition, we consider the scenario where only one user is jammed to better understand the impact of partial model poisoning on global SFL outcomes. 
Finally, we also investigate the role of the jammer by analyzing the impact of a simple barrage jammer compared to the worst-case jammer. 

\subsubsection{Sequence Classification}
\begin{figure*}[htbp]
    \centering
    \includegraphics[width=1\textwidth]{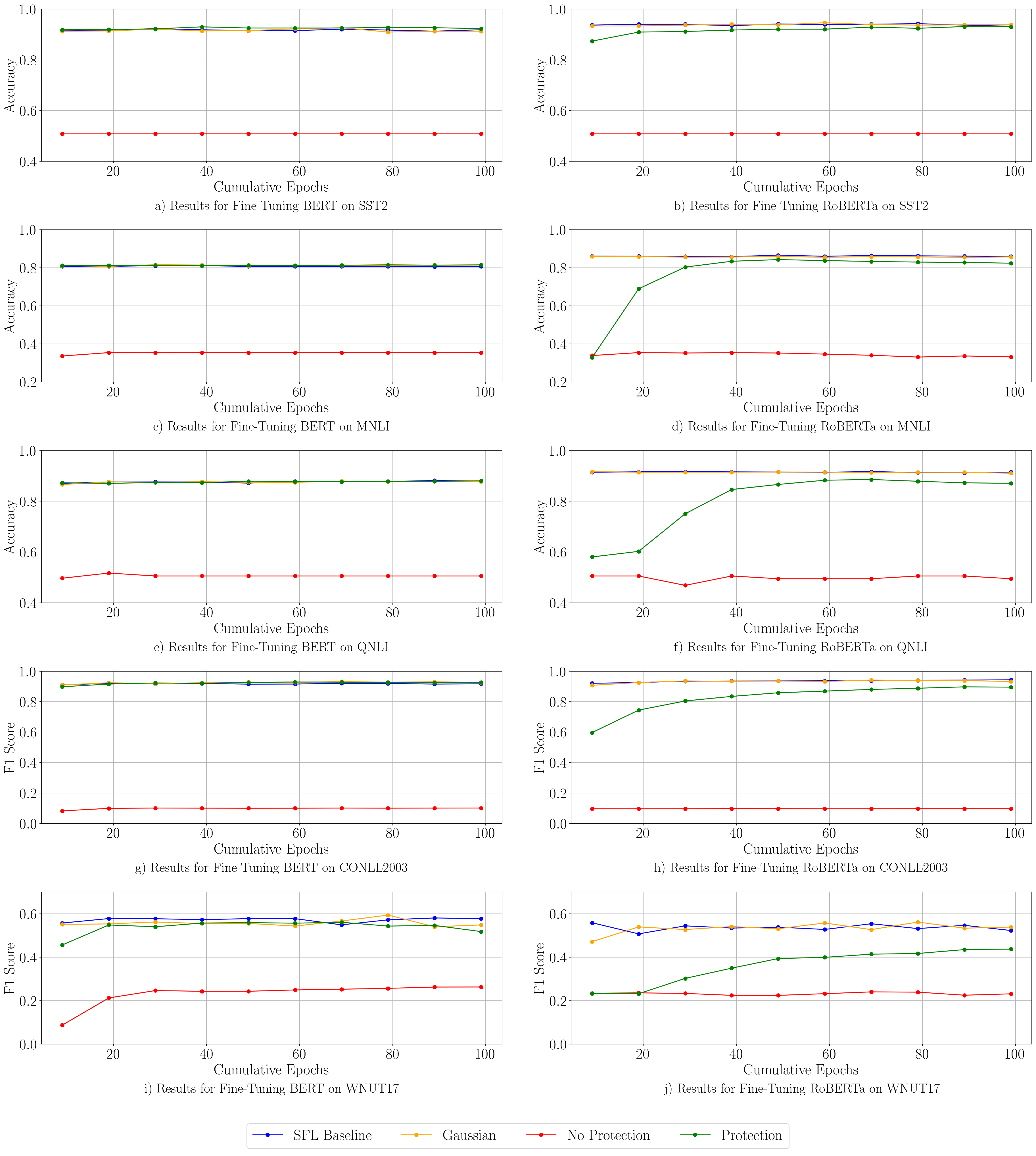}
    \caption{Global SFL model performance plots for NLP experiments with BERT/RoBERTa, evaluated after each of the $N_\text{rounds}$ global rounds for all four scenarios (\emph{SFL Baseline}, \emph{Gaussian}, \emph{No Protection}, \emph{Protection}) using Accuracy and F1 Scores (higher is better). 
    Accuracy is calculated as the ratio of correctly classified sentences/words to the total number of instances, while the F1 Score is the harmonic mean of precision (ratio of true positive observations to the total number of predicted positives) and recall (ratio of true positive observations to the number of actual positives, i.e. the sum of true positives and false negatives).}
    \label{fig:jointplot}
\end{figure*}

As shown in Table \ref{tab:sfl_results}, for all three SC datasets, R-SFLLM is able to consistently safeguard the distributed training in general, leading to resilient global models with classification accuracies near-identical or very close to the baselines. 
%
For instance, BERT achieves near-identical performance to the SST2 baseline with accuracies above $91\%$ for the scenarios when worst-case jamming is absent (\textit{Gaussian}) and when R-SFLLM protection against it is employed (\textit{Protection}). 
This indicates that AWGN alone does not significantly impact the embeddings and is negligible within a statistical variance, thereby acting as a light regularizer. 
As for the scenario where no protection is provided (\textit{No Protection}), the worst-case jammer is successful in maximally disrupting the distributed training, resulting in a global model accuracy of around $50\%$. 
Thus, the global SFL model is no better than simple guessing, in other words, it has not been able to learn anything and subsequently has worst-case binary classification performance. 
This answers the question of whether worst-case jamming translates into worst-case performance positively. 
Moreover, each client observes near optimal performance from early epochs on, such that the global model already converges after the first global round, shown in \refig{fig:jointplot}\textcolor{blue}{a}. 
The same can be observed for MNLI and QNLI in \refig{fig:jointplot}\textcolor{blue}{c} and \refig{fig:jointplot}\textcolor{blue}{d}. 
This demonstrates that BERT is well pre-trained such that SC is an easy fine-tuning task.

In comparison, RoBERTa achieves a slightly better overall baseline performance of around $93\%$ for SST2, indicating a potential advantage due to its more complex architecture. 
However, for the scenario \textit{Protection}, \refig{fig:jointplot}\textcolor{blue}{b} shows that the global model starts off with a slightly lower accuracy and converges to the baseline after the second global round. 
While negligible in this case, this indicates that RoBERTa is more sensitive to noisy embeddings. This is due to the fact that jamming cannot be mitigated perfectly, resulting in higher MSEs for scenario \textit{Protection} as compared to \textit{Gaussian}. 
In particular, RoBERTa does not use segment embeddings and relies solely on position and token embeddings, unlike BERT, which uses all three embedding types. 
Furthermore, RoBERTa's pre-training process involves more extensive and diverse data, leading to a model that is more finely tuned to the nuances of language. 
This renders RoBERTa more susceptible to perturbations in embeddings, as it has learned to rely on subtle features that can be disrupted by noise. 
This noise sensitivity is more pronounced for the QNLI and partly for the larger MNLI dataset, shown in \refig{fig:jointplot}\textcolor{blue}{f} and \refig{fig:jointplot}\textcolor{blue}{d}, respectively, such that RoBERTa converges to the QNLI baseline only after the sixth global round. 

To investigate the fairness of our resource allocation strategy, \refig{fig:qnli_all_clients} shows the QNLI performance for RoBERTa clients 1 to 3, respectively, where clients 1 and 2 approach the baseline whereas client 3 aligns more to the observed global SFL model. 
When further comparing the MSEs for each user in Table \ref{tab:sfl_results}, client 1 experiences the lowest MSE, while client 3 experiences the highest MSE, which is about 2.8 times higher than the one from client 1. 
This indicates that the developed resource allocation of the proposed protection scheme is \emph{not fair}. 
However, this difference, while significant, does not seem to particularly affect the BERT model. 
This corroborates that RoBERTa is more sensitive to noisy embeddings in general. 
However, since this is not necessarily observed for RoBERTa on the SST2 dataset, there likely is an additional dependence on the data distribution, such that a potentially non-independently-and-identically distributed (non-IID) data split may further decrease the performance in case of corrupted embeddings. 
Thus, fine-tuning RoBERTa may result in an initially worse model if some clients underperform. 

Nevertheless, R-SFLLM remains successful in mitigating the worst-case jammer, albeit after a few global rounds. 
This shows that RoBERTa makes use of the adversarial noise and robustifies over the training period, thus benefitting from the \emph{adversarial training character} of R-SFLLM, in which the additional noise, if moderate, helps the model to regularize over time, thereby managing to yield a close-to-optimal global model performance after a few global rounds. 
This is similar to traditional adversarial training \cite{adversarial_training_per_layer}, however targets only embeddings and no other layers instead. 

\begin{table}[t]
    \centering
    \resizebox{0.5\textwidth}{!}{%
    \begin{tabular}{llcccc}
        \toprule
        & \textbf{Dataset / Model / MSE} & \textbf{SFL Baseline} & \textbf{Gaussian} & \textbf{Protection} & \textbf{No Protection} \\
        \midrule

        \multirow{14}{*}{\rotatebox{90}{\textbf{NLP Results}}}
        & SST2 (BERT) & 91.9\% & 91.2\% & 92.3\% & 50.9\% \\
        & SST2 (RoBERTa) & 93.2\% & 93.8\% & 93.0\% & 50.9\% \\
        \cmidrule(lr){2-6}
        & QNLI (BERT) & 87.9\% & 87.8\% & 88.0\% & 50.5\% \\
        & QNLI (RoBERTa) & 91.5\% & 91.0\% & 87.1\% & 49.5\% \\
        \cmidrule(lr){2-6}
        & MNLI (BERT) & 80.8\% & 81.6\% & 81.5\% & 35.4\% \\
        & MNLI (RoBERTa) & 86.0\% & 85.7\% & 82.5\% & 33.2\% \\
        \cmidrule(lr){2-6}
        & CONLL2003 (BERT) & 92.2\% & 92.5\% & 92.7\% & 10.1\% \\
        & CONLL2003 (RoBERTa) & 93.6\% & 93.2\% & 89.5\% & 9.7\% \\
        \cmidrule(lr){2-6}
        & WNUT17 (BERT) & 58.5\% & 54.8\% & 51.7\% & 26.3\% \\
        & WNUT17 (RoBERTa) & 54.1\% & 53.8\% & 43.7\% & 23.1\% \\
        \toprule

        \multirow{8}{*}{\rotatebox{90}{\textbf{CV Results}}}
        & MNIST & 98.5\% & 98.0\% & 97.5\% & 9.8\% \\
        & Cars & 86.3\% & 85.3\% & 84.1\% & 7.5\% \\
        & DTD & 81.5\% & 80.4\% & 78.6\% & 6.7\% \\
        & EuroSAT & 97.7\% & 97.2\% & 96.7\% & 10.8\% \\
        & GTSRB & 94.3\% & 93.8\% & 93.5\% & 8.5\% \\
        & RESISC45 & 96.5\% & 96.0\% & 95.5\% & 5.3\% \\
        & SUN397 & 78.2\% & 77.7\% & 76.9\% & 10.2\% \\
        & SVHN & 93.1\% & 92.6\% & 92.1\% & 7.7\% \\
        \toprule
        
        \multirow{3}{*}{\rotatebox{90}{\textbf{MSE}}}
        & Client 1 MSE & $-\infty$ dB & -10.2 dB & -7.9 dB & 23.3 dB \\
        & Client 2 MSE & $-\infty$ dB & -9.8 dB & -5.9 dB & 25.9 dB \\
        & Client 3 MSE & $-\infty$ dB & -10.5 dB & -3.4 dB & 27.1 dB \\
        \bottomrule
    \end{tabular}
    }
    \caption{Fine-Tuning results for NLP and CV experiments. NLP: BERT and RoBERTa LLMs on SST2, QNLI, MNLI, CONLL2003, and WNUT17 datasets across different SFL training scenarios, evaluated via Accuracy (for SC) and F1 Scores (for NER). CV: CLIP ViT-B/16 VLM on MNIST, Cars, DTD, EuroSAT, GTSRB, RESISC45, SUN397, and SVHN image classification datasets, evaluated via Accuracy.}
    \label{tab:sfl_results}
    \vspace{-0.5cm}
\end{table}

\subsubsection{Named Entity Recognition}

Table \ref{tab:sfl_results} shows that BERT similarly achieves near-identical performance to the SFL baseline for the smaller CONLL2003 dataset with F1 scores above $92\%$ for the scenarios \textit{Gaussian} and \textit{Protection}. 
Again, \textit{No Protection} results in maximal disruption, with consistent F1 scores around $10\%$. This suggests that the model is either missing almost all of the entities (low recall) or predicting almost all entities incorrectly (low precision), which ultimately renders the obtained global model unusable for NER. 
Thus, worst-case jamming results in worst-case model performance. 
In addition, \refig{fig:jointplot}\textcolor{blue}{g} again shows that each client observes good performance from early epochs on. 
This further corroborates that the BERT architecture is robust against noisy embeddings, even for unfair resource allocation in wireless SFL setups. 

RoBERTa achieves similar outcomes for CONLL2003, matching the baseline performance with F1 scores of around $93\%$. 
Worst-case global model performance is observed for scenario \textit{No Protection} with F1 scores around $10\%$ as well. 
However, as observed for SC, the global SFL model for scenario \textit{Protection} starts out at lower F1 scores of around $60\%$ and converges gradually to the baseline toward the end of the global training, shown in \refig{fig:jointplot}\textcolor{blue}{h}. 
This can be attributed to the same explanation as before, such that RoBERTa is more vulnerable to noisy embeddings, particularly when some clients are subjected to higher MSEs due to unfair resource allocation. 
Thus, similarly as for SC, the global SFL performance converges to the baseline after a few global rounds, where RoBERTa makes use of R-SFLLM's adversarial training character and robustifies over the fine-tuning period.

In contrast, WNUT17 represents a more challenging dataset, containing viral social media posts that are highly informal and complex to categorize for LLMs. 
Thus, baseline performance is significantly lower. 
In addition, WNUT17 consists of only few samples, such that each client is challenged with poorer insights into the data.
However, even in this case, BERT achieves good but not great performance for all relevant scenarios given the small dataset. 
Again, the worst-case jammer is able to successfully impair the global model with F1 scores of around $25\%$, which corresponds to the performance of the baseline after only one epoch, shown in \refig{fig:jointplot}\textcolor{blue}{i}. 
Hence, the global model continues to miss almost all of the entities, having not progressed at all. 
Furthermore, RoBERTa's sensitivity to noisy embeddings due to unfair resource allocation becomes more pronounced when dealing with particularly small datasets, such as in this case. 
Specifically, \refig{fig:wnut_all_clients} shows that while clients 1 and 2 attain the baseline, client 3 with higher MSE does not. 
This results in a worse performing global model, shown in \refig{fig:jointplot}\textcolor{blue}{j}, where the projected performance trajectory indicates that RoBERTa not only needs more data but also more global SFL rounds to successfully mitigate the worst-case jamming impact. 
However, the performance after 10 global rounds is already significantly better than in the case without R-SFLLM protection.
These results suggest that both BERT and RoBERTa become more susceptible to noisy embeddings in wireless SFL when small datasets with potentially non-representative samples are used for training. 
The model might thus not generalize well nor benefit from the adversarial training character of R-SFLLM. 

\subsubsection{Jamming only one User}
\begin{figure}[htbp]
    \centering
    \includegraphics[width=0.47\textwidth]{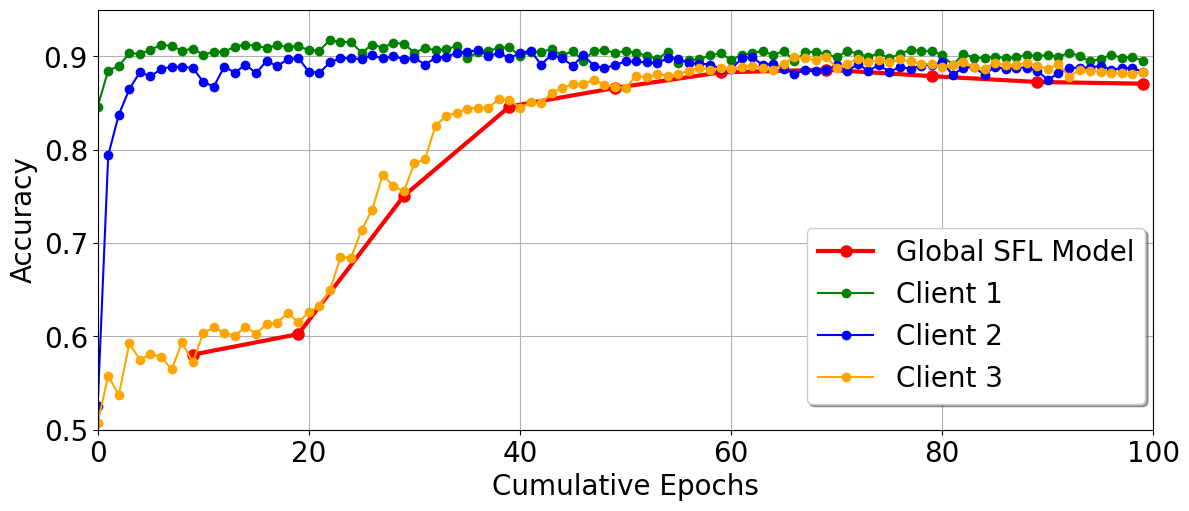}
    \caption{Results for fine-tuning RoBERTa on QNLI with all client plots.}
    \label{fig:qnli_all_clients}
\end{figure}

\begin{figure}[htbp]
    \centering
    \includegraphics[width=0.47\textwidth]{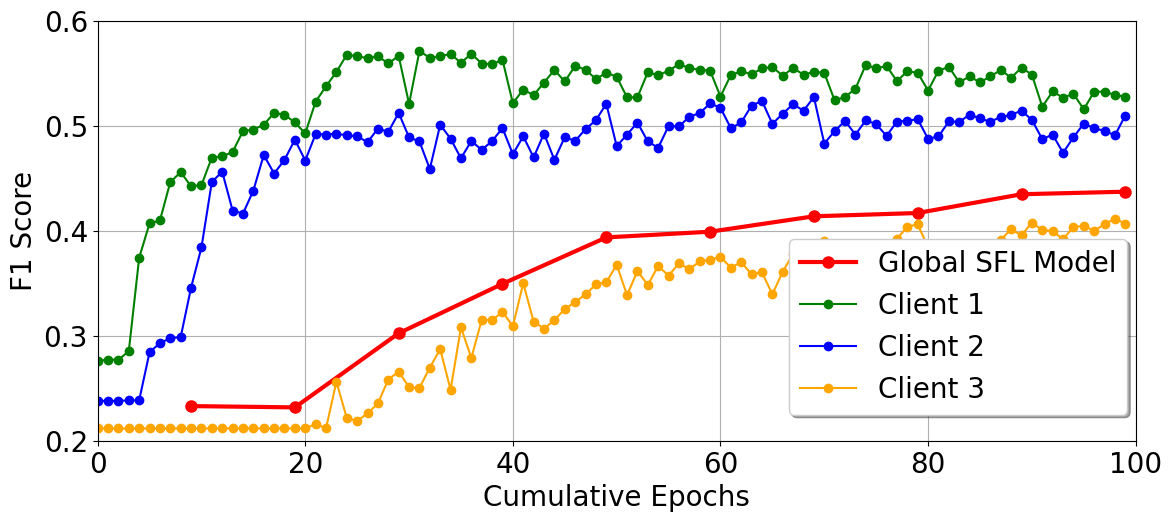}
    \caption{Results for fine-tuning RoBERTa on WNUT17 with all client plots.}
    \label{fig:wnut_all_clients}
\end{figure}

\begin{figure}[htbp]
    \centering
    \includegraphics[width=0.47\textwidth]{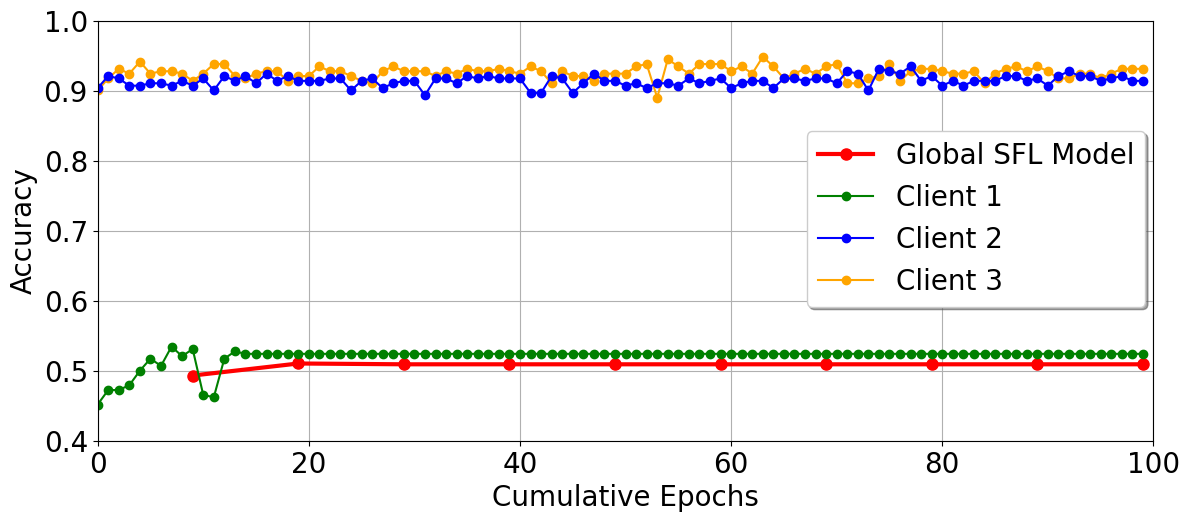}
    \caption{Results for fine-tuning BERT on SST2 when jamming only client 1. The global SFL model cannot recover even if only one party is targeted.}
    \label{fig:one_jammer_only}
\end{figure}

Based on the previous discussions and the results in \refig{fig:qnli_all_clients} and \refig{fig:wnut_all_clients}, it suffices if only one SFL party performs suboptimal for the global model to decrease in performance. 
However, this was investigated under the setting that all SFL clients are jammed simultaneously and where always client 3 suffered from higher MSEs due to unfair resource allocation as a result of our water-filling approach. 
To extend the investigation, \refig{fig:one_jammer_only} provides simulation results for BERT on SST2, where only one party (client 1) is jammed, and shows that SFL is highly sensitive to aggregating even only one (randomly) poisoned model, regardless of whether BERT is more robust against noisy embeddings. 
Note that similar worst-case results can be obtained for RoBERTa, other datasets, and other clients. This further corroborates the findings in \cite{yoo2022backdoor}, where it was shown that few poisoned clients suffice to compromise the global model. 
Our results show that this can be achieved via jamming attacks in wireless SFL.

\subsubsection{Barrage Jamming}
\begin{figure}[htbp]
    \centering
    \includegraphics[width=0.47\textwidth]{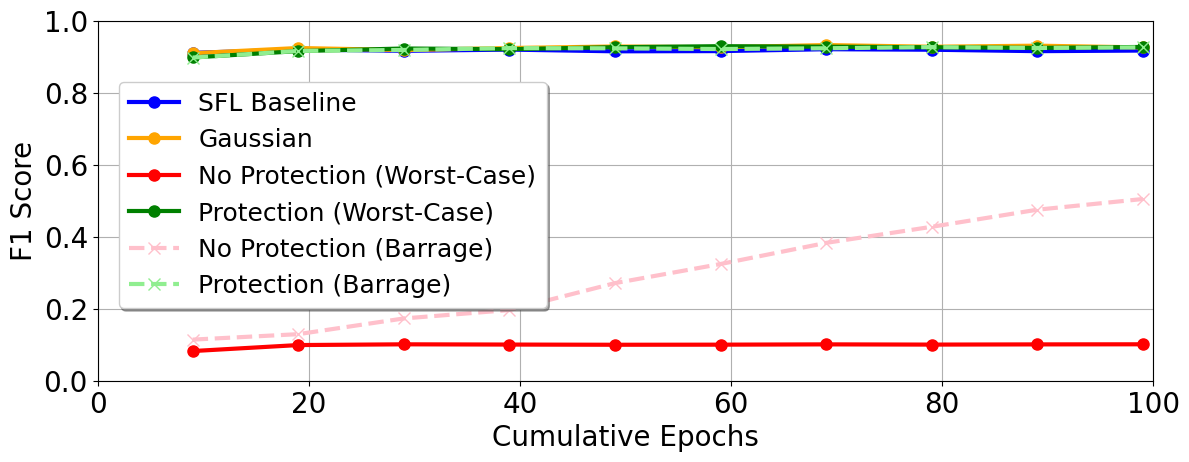}
    \caption{Results for fine-tuning BERT on CONLL2003 with additional barrage jamming. The worst-case jammer is consistently stronger over all epochs and global SFL rounds while the barrage jammer improves gradually over time.}
    \label{fig:bert_barrage}
\end{figure}

\begin{figure}[htbp]
    \centering
    \includegraphics[width=0.47\textwidth]{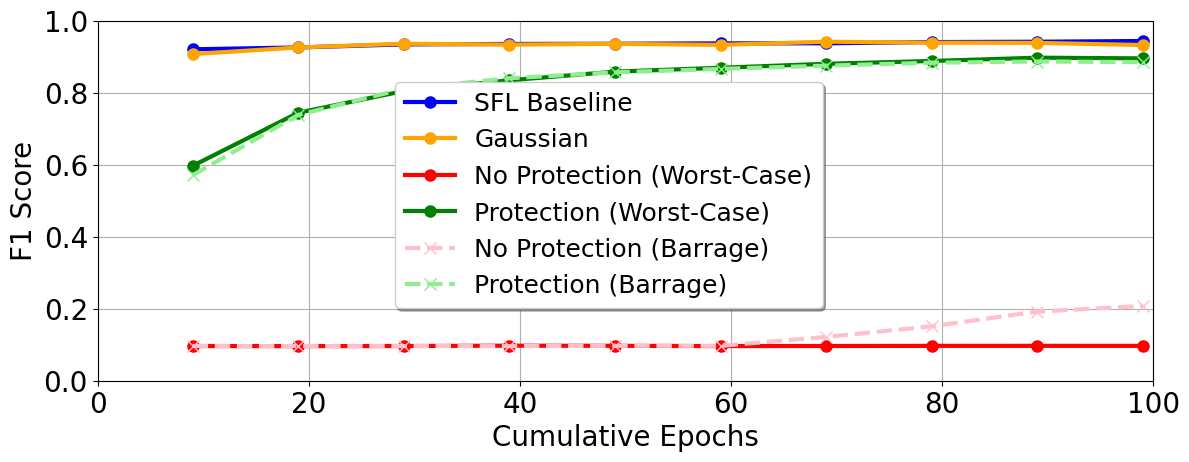}
    \caption{Results for fine-tuning RoBERTa on CONLL2003 with additional barrage jamming. The worst-case jammer is consistently stronger over all epochs and global SFL rounds while the barrage jammer improves only marginally after the sixth round.}
    \label{fig:roberta_barrage}
\end{figure}

Next, we quantify how adversaries other than worst-case jammers affect the SFL performance. 
To this end, we employ a barrage jammer with covariance $\boldsymbol{C}_{\boldsymbol{u}_{nk}} = P_J / (N_J N K)$ and demonstrate results for BERT and RoBERTa on CONLL2003 in \refig{fig:bert_barrage} and \refig{fig:roberta_barrage}, respectively. 
In both cases, barrage jamming has a non-negligible impact on the performance for scenario \textit{No Protection}, which at first resembles the worst-case but improves with later global rounds. 
For BERT, SFL performance gradually increases with every global round and achieves an F1 score of $51\%$, which is still $41\%$ less than the baseline. 
For RoBERTa, performance increases only marginally after the sixth round and achieves an F1 score of $21\%$, being only $11\%$ higher than the worst-case. 
This is interesting as clients 1,2, and 3 observe MSEs of 5.3dB, 8.0dB, and 9.2dB, respectively, which are significantly lower than for the worst-case, but still higher than for scenario \textit{Protection}. 
This shows that small noise acts as a regularizer but higher noise beyond a certain threshold results in severe corruption, particularly for noise-sensitive RoBERTa models. 

\vspace{-0.5cm}

\vspace{-0.5cm}

\vspace{-0.5cm}

\vspace{-0.5cm}

\noindent
Nevertheless, only the worst-case jammer is able to consistently maintain its detrimental performance. 
As for scenario \textit{Protection}, barrage jamming matches the previous experiments. 
Thus, R-SFLLM mitigates adversarial jamming with consistent performance, regardless of whether worst-case, barrage, or any other jammer is employed, demonstrating its versatility across most practical jamming scenarios.
\vspace{-1em}

\subsection{R-SFLLM Simulation Results for VLM Models.}
\label{sec:simulation_results_VLM}

\begin{figure}[t]
    \centering
    \includegraphics[width=1\linewidth]{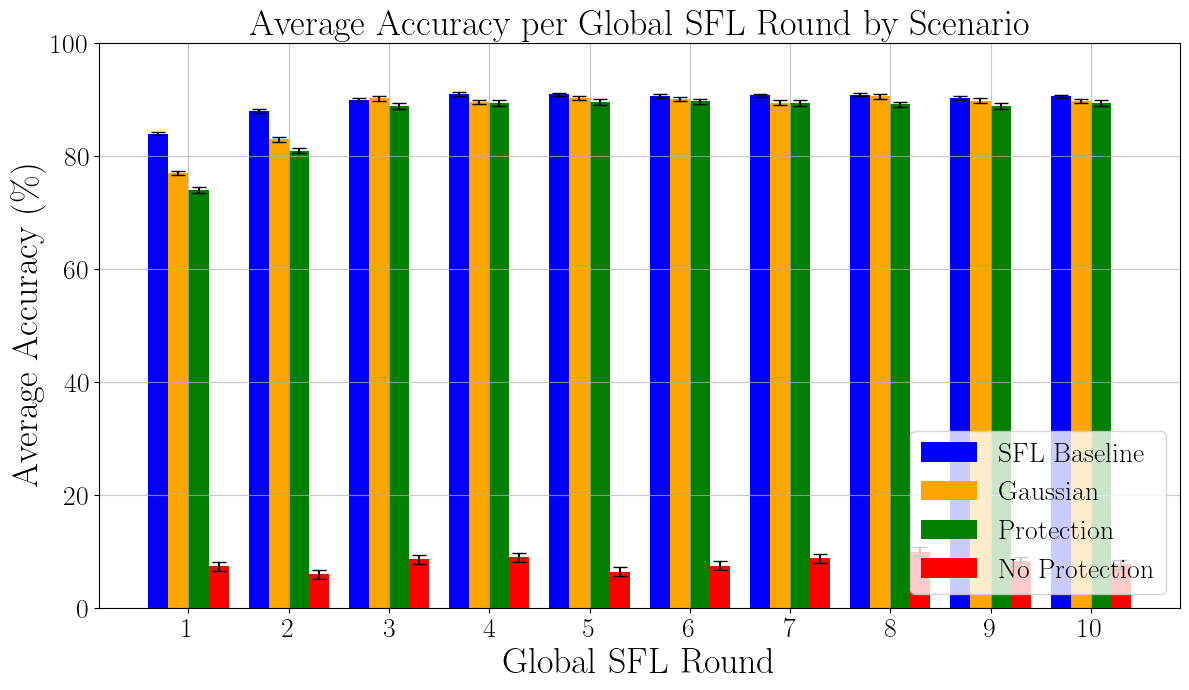}
    \caption{Global SFL model performance for CV experiments with CLIP ViT-B/16, evaluated after each of the $N_\text{rounds}$ global SFL rounds and averaged across all datasets. R-SFLLM effectively safeguards multi-modal embeddings, achieving performance close to the SFL baseline after the third global round.}
    \label{fig:vlm_plot}
    \vspace{-1em}
\end{figure}

Compared to BERT and RoBERTa, VLMs are multi-modal models that process both images and text in parallel.
In particular, OpenAI's CLIP ViT-B/16 VLM considered here uses a vision transformer (ViT) backbone, paired with a text transformer. 
Both encoders produce embeddings that are contrasted during training, i.e., matching image and text embeddings are pulled closer together while non-matching pairs are pushed apart, yielding multi-modal embeddings.
This makes it particularly interesting to study the impact of adversarial noise on both language and image understanding. 
Furthermore, since a larger volume of corresponding embeddings must be transmitted during SFL, the system might become more susceptible to adversarial attacks, impacting resilience more severely than in lower-bandwidth NLP use cases.
To investigate this, we fine-tune the CLIP VLM on eight diverse image classification datasets, including samples ranging from European satellite imagery in EuroSAT to German traffic signs in GTSRB.
Results for all scenarios are shown in Table \ref{tab:sfl_results}, and \refig{fig:vlm_plot} illustrates the model performance across global SFL rounds, averaged over all considered datasets.

In general, the \emph{SFL baseline} converges after the third global round, achieving an average accuracy of 91\%. 
This is partly due to variation in sample sizes, where smaller datasets, such as DTD and Cars, require more global rounds for convergence, while training on larger datasets like SVHN and MNIST typically converges after the first round. 
A similar trend is observed for scenarios \emph{Gaussian} and \emph{Protection}, both of which also converge by the third global round.
While convergence is observed, the results clearly show the pronounced impact of noise for the more vulnerable multi-modal embeddings of the VLM architecture.
Since both text and image embeddings are exposed to noise, the contrastive learning objective of VLMs is more strongly affected, leading to mismatching text labels for images, and thus to incorrect classification outcomes.
This results in an initially larger performance gap between the SFL baseline and scenarios \emph{Gaussian} and \emph{Protection}.
For example, after the first global round, the accuracy of \emph{Gaussian} (\emph{Protection}) is around 77\% (74\%), thereby 7\% (10\%) less than the SFL baseline.
As training progresses, this performance gap narrows significantly, in particular, dropping to less than 1\% after the third global round.
These results highlight two key observations: First, the VLM's multi-modal embeddings are indeed more susceptible to both benign (i.e., Gaussian) and adversarial noise during the early rounds of training. 
Second, R-SFLLM is able to maintain performance comparable to \emph{Gaussian}, thus achieving close-to-baseline accuracies toward the end of training.
This underscores R-SFLLM's effectiveness in mitigating worst-case adversarial jamming across a variety of datasets, where \emph{No Protection} would otherwise result in average performance as low as 8\%, reflecting the model's inability to correctly match images and text labels.
Thus, across all classification tasks, R-SFLLM achieves performance close to the SFL baseline by the third global round, by which point the VLM has adapted to the adversarial training signal.
Consistent with our previous NLP results, this holds even for the smallest datasets, such as DTD (4k samples) and Cars (8k samples), where R-SFLLM achieves accuracies within 2–3\% of the baseline (see Table \ref{tab:sfl_results}).
Furthermore, larger datasets such as MNIST remain highly competitive with 97.5\% accuracy as more samples are observed during training.
Thus, R-SFLLM enables the VLM to effectively learn contrastive embeddings even in the presence of adversarial jamming, resulting in strong performance across diverse datasets with varying sizes and distributions.
This demonstrates that R-SFLLM is applicable across a range of transformer-based language models, including more complex multi-modal architectures.

\vspace{-1em}

\subsection{Ablation Results and Sensitivity Analysis.}
\label{sec:ablations}

We examine the impact of varying the number of users $Q$ and jamming antennas $N_J$ on the performance of R-SFLLM, demonstrating that our framework is both scalable and effective for various adversarial SFL setups.  

\subsubsection{Increasing the Number of Users $Q$}
\label{sec:ablations_users}

\begin{figure}[htbp]
    \centering
    \includegraphics[width=1\linewidth]{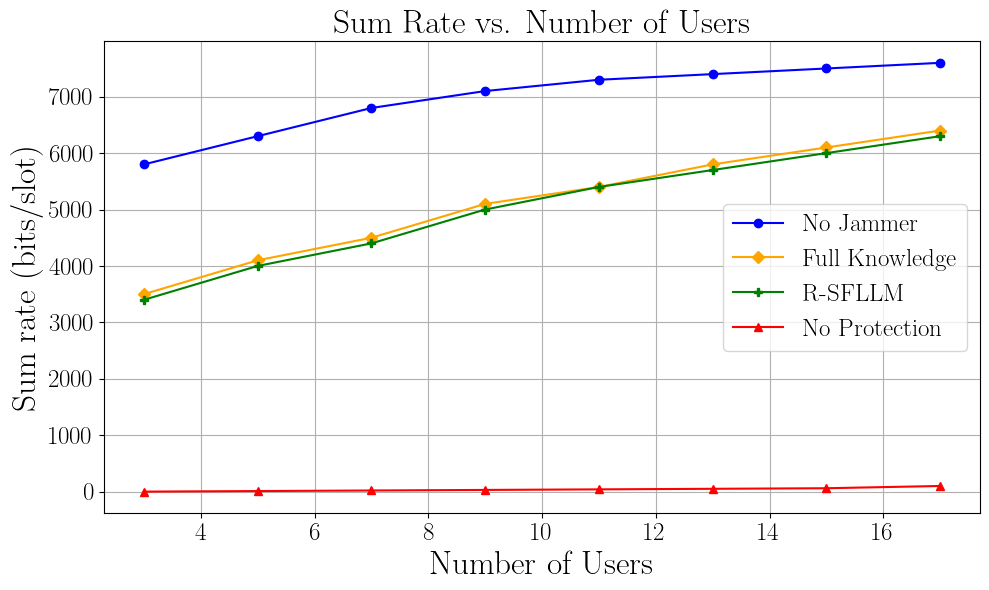}
    \caption{Sum rate vs. number of users $Q$ for scenarios No Protection, No Jammer, R-SFLLM, and Full Knowledge. R-SFLLM approximates the optimal solution with Full Knowledge and increases the sum rate for increasing $Q$.}
    \label{fig:ablation_users}
\end{figure}

\begin{figure}[htbp]
    \centering
    \includegraphics[width=1\linewidth]{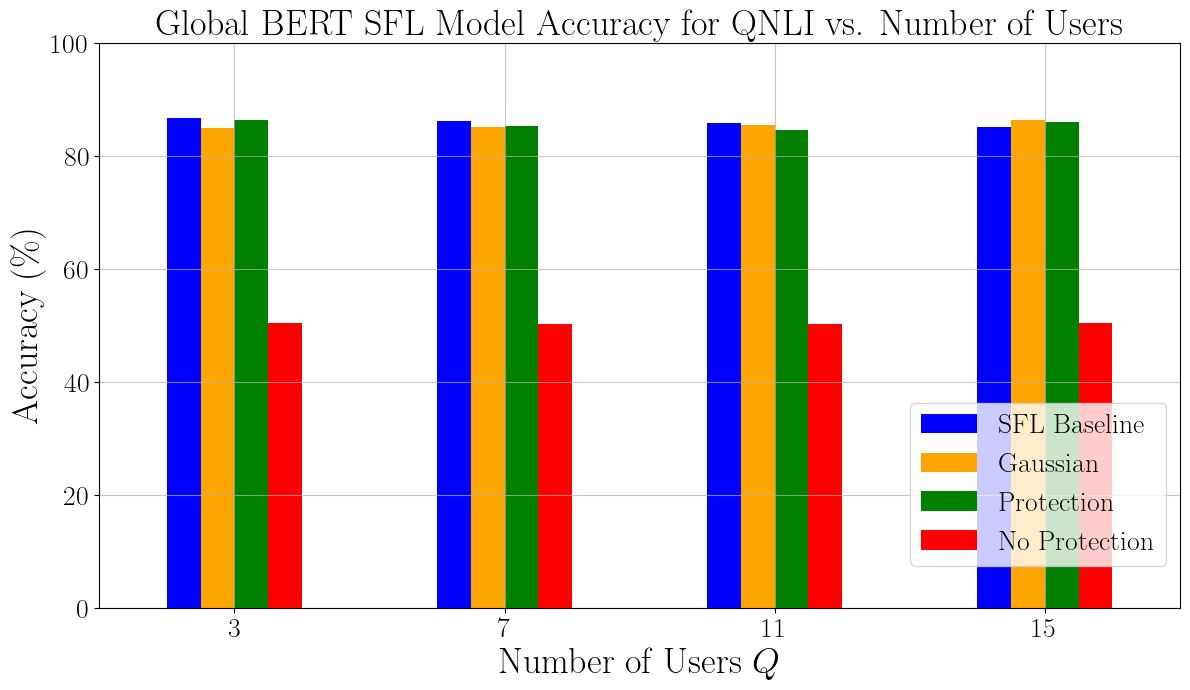}
    \caption{Global BERT SFL model performance for varying users $Q$. R-SFLLM remains effective and scalable when the number of SFL participants increases.}
    \label{fig:ablation_bert_qnli}
\end{figure}

In \refig{fig:ablation_users}, we compare sum rates for the scenarios \emph{No Protection}, \emph{No Jammer}, \emph{R-SFLLM}, and \emph{Full Knowledge}, where the latter represents an arbitrary, but optimal reference anti-jamming strategy that requires full knowledge about the jamming covariance matrix $\boldsymbol{C}_{\boldsymbol{u}_{nk}}$, as studied in \cite{andrei2024resilientbydesign}.
We see that, in general, the sum rate increases with increasing number of users, and that our R-SFLLM framework approximates the sum rate for the optimal reference solution.
In contrast, the worst-case jammer manages to bring down the sum rate to almost zero for scenario \emph{No Protection}. 
This demonstrates that our proposed sensing-assisted anti-jamming strategy is highly competitive compared to established algorithms that require full system and jammer knowledge. 
To validate whether this translates directly into SFL performance, we fine-tune BERT on the larger QNLI binary classification dataset for SC with $Q = \{3, 7, 11, 15\}$ users, respectively, and show corresponding performance plots in \refig{fig:ablation_bert_qnli}.
In particular, we see that performance is sustained when increasing the number of SFL participants, such that outcomes only differ by negligible statistical variance around 1\%.
This demonstrates that R-SFLLM is highly scalable.

\subsubsection{Increasing the Number of Jamming Antennas $N_J$}
\label{sec:ablations_jamming_antennas}

\begin{figure}[htbp]
    \centering
    \includegraphics[width=1\linewidth]{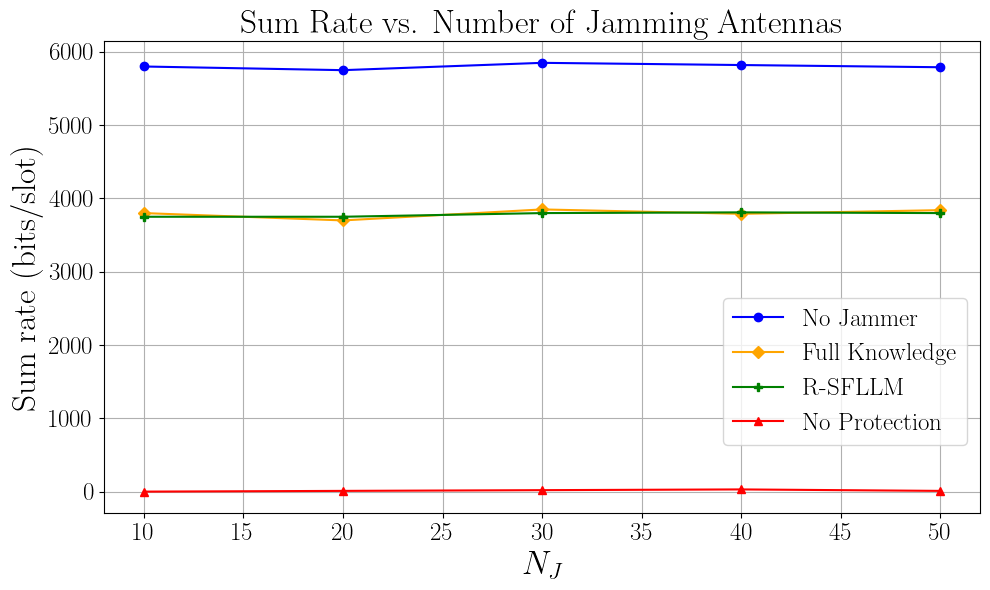}
    \caption{Sum rate vs. number of jamming antennas $N_J$ for scenarios No Protection, No Jammer, R-SFLLM, and Full Knowledge. R-SFLLM approximates the optimal reference solution and remains resilient against higher $N_J$.}
    \label{fig:ablation_antennas}
\end{figure}

In \refig{fig:ablation_antennas}, we compare sum rates across the same scenarios when equipping the adversarial jammer with more antennas $N_J$.
We show that R-SFLLM maintains consistent performance, achieving rates close to the optimal reference solution and demonstrating that our proposed framework is resilient even against jammers with very large numbers of antennas.
To extend this discussion, we further examined R-SFLLM's anti-jamming performance across a wide range of jamming DoAs, power budgets, and scale parameters $\eta$ in our previous work in \cite{andrei2024resilientbydesign}.
There, we showed that anti-jamming remains effective and scalable, independent of the number of wireless users or jammer capabilities, when $\eta$ is chosen to be significantly larger than the noise level $\sigma^2$, a practical heuristic that we carefully validated and discussed in more detail in \refsec{sec:role_of_doa}.

\vspace{-1em}

\subsection{Summary of Results.}
\label{sec:summary}

In our experiments, we showed that jamming LLM embeddings during SFL leads to severe performance degradation, with worst-case jamming resulting in worst-case training performance.
Our investigation further revealed that some models are particularly sensitive to noisy embeddings. 
For example, RoBERTa is more susceptible due to the lack of stabilizing segment embeddings, while noise during CLIP VLM's contrastive optimization challenges the learning of multi-modal embeddings.
However, across all settings, our proposed R-SFLLM framework was able to effectively safeguard SFL training by mitigating noise from adversarial jammers, including worst-case ones, thereby ensuring resilience and achieving performance near identical to SFL baselines.
To ensure wide-range applicability, we evaluated R-SFLLM for both NLP and CV domains using three different models (BERT, RoBERTa, and multi-modal CLIP ViT-B/16) across 13 different datasets of varying sizes and distributions, showing that R-SFLLM generalizes well to different types of transformer-based architectures in practice.
In extensive ablation studies, we demonstrated that R-SFLLM is scalable and can be applied to large-scale SFL settings, while providing resilience against various types of adversarial jammers, regardless of their power, number of antennas, or system knowledge. 
Furthermore, we benchmarked R-SFLLM against an optimal reference solution, showing closely matching outcomes and thereby verifying its competitiveness against alternative anti-jamming strategies.

\section{Conclusion}
In this paper, we investigated the problem of adversarial jamming attacks in wireless SFL with language models. 
We demonstrated both theoretically and experimentally that jamming critical embedding parameters leads to significantly worse model outcomes. 
Specifically, we showed that worst-case jamming results in worst-case performance, regardless of how many clients are being targeted. 
To address this, we proposed R-SFLLM, a scalable, sensing-assisted anti-jamming framework that leverages information about the adversary's DoAs to integrate resilience directly into the wireless system by design. 
We justified this wireless approach to resilience by showing that the LLM model error is upper bounded by the communication MSE under a relaxed $(\boldsymbol{L}_0, \boldsymbol{L}_1)$-smoothness assumption, thereby emphasizing that the physical layer directly impacts the training. 
Extensive experiments in both NLP and CV domains, using three different model architectures and 13 datasets, validate the effectiveness of R-SFLLM and demonstrate the critical need to safeguard embedding parameters in SFL against adversarial jamming attacks to ensure model integrity in distributed training over wireless networks.

\vspace{-0.5cm}

\vspace{-0.5cm}

\hfill 


\appendices

\section{Proof of Lemma 2}
\label{app:A}
First, we adapt the original proof for Lemma \ref{L1} in \cite{robustness} and reformulate its initial statement toward the loss divergence: 
\begin{align}
    L(\boldsymbol{y}) &- L(\boldsymbol{x}) =  \nabla L(\boldsymbol{x})^T (\boldsymbol{y} - \boldsymbol{x}) \nonumber \\ 
    &+ \int_{0}^1 \langle \nabla L(\boldsymbol{x} + u(\boldsymbol{y} - \boldsymbol{x})) - \nabla L(\boldsymbol{x}), \boldsymbol{y} - \boldsymbol{x} \rangle du.
\end{align}

Second, applying the norm $|.|$ and triangle inequality yields 
\begin{align}
    | L(&\boldsymbol{y}) - L(\boldsymbol{x}) | \leq  \left | \nabla L(\boldsymbol{x})^T (\boldsymbol{y} - \boldsymbol{x}) \right | \nonumber \\ 
    &+ \left | \int_{0}^1 \langle \nabla L(\boldsymbol{x} + u(\boldsymbol{y} - \boldsymbol{x})) - \nabla L(\boldsymbol{x}), \boldsymbol{y} - \boldsymbol{x} \rangle du \right |. \label{eq:app_a_integral}
\end{align}

Third, we use the upper bound on this integral from the proof in \cite{robustness} to further obtain
\begin{align}
    | &L(\boldsymbol{y}) - L(\boldsymbol{x}) | \leq  \left | \nabla L(\boldsymbol{x})^T (\boldsymbol{y} - \boldsymbol{x}) \right | \nonumber \\ 
    &+ \left | \sum_{j=1}^E \left [  
    L_{0,j} + L_{1,j} \left | \frac{\partial L (\boldsymbol{x}) }{\partial x_j} \right | \right ] |y_j - x_j| \cdot || \boldsymbol{y} - \boldsymbol{x}||_2 \right |.
\end{align}

By defining the vectors $\boldsymbol{u} = [u_1 \hdots u_E]^T$ with $u_j = L_{0,j} + L_{1,j} \left | \frac{\partial L (\boldsymbol{x})}{\partial x_j} \right |$ and $\boldsymbol{v} = [v_1 \hdots v_E]^T$ with $v_j = |y_j - x_j|$, above expression can be reformulated using scalar products, i.e.
\begin{align}
    \left | L(\boldsymbol{y}) - L(\boldsymbol{x}) \right | &\leq \left | \nabla L(\boldsymbol{x})^T (\boldsymbol{y} - \boldsymbol{x}) \right| + \left | \boldsymbol{u}^T \boldsymbol{v} \cdot \| \boldsymbol{y} - \boldsymbol{x} \|_2 \right |.
\end{align}

Next, applying the Cauchy-Schwarz inequality on the right-hand side and identifying $\| \boldsymbol{v} \|_2 = \| \boldsymbol{y} - \boldsymbol{x} \|_2$ yields
\begin{align}
    \left | L(\boldsymbol{y}) - L(\boldsymbol{x}) \right | &\leq \| \nabla L(\boldsymbol{x}) \|_2 \cdot \| \boldsymbol{y} - \boldsymbol{x} \|_2 \nonumber \\ &+ \| \boldsymbol{u} \|_2 \cdot \| \boldsymbol{y} - \boldsymbol{x} \|_2^2.
\end{align}

With $\boldsymbol{u} = \boldsymbol{L_0} + \boldsymbol{L_1} \odot | \nabla L(\boldsymbol{x}) |$, where $\odot$ denotes the Hadamard product and $\nabla  L(\boldsymbol{x})$ is the gradient vector of $L$, the statement in Lemma \ref{P0} is obtained.
\section{Proof of Proposition 2}
\label{app:B}
Plugging in $\boldsymbol{e}_q, \hat{\boldsymbol{e}}_q$ into the upper bound on the loss divergence in \eqref{eq:upper_bound_losses_without_tau_lemma_2} and taking the gradient w.r.t. $\boldsymbol{e}_q$ $(\nabla_{\boldsymbol{e}_q})$ gives 
\begin{align}
    \left | L(\boldsymbol{e}_q) \right. &- \left. L(\hat{\boldsymbol{e}}_q) \right | \leq \| \nabla_{\boldsymbol{e}_q} L(\boldsymbol{e}_q) \|_2 \cdot \| \boldsymbol{e}_q - \hat{\boldsymbol{e}}_q \|_2 \nonumber \\ 
    &+ \left \| \boldsymbol{L_0} + \boldsymbol{L_1} \odot | \nabla_{\boldsymbol{e}_q} L(\boldsymbol{e}_q) | \right \|_2 \cdot \| \boldsymbol{e}_q - \hat{\boldsymbol{e}}_q \|_2^2.
\end{align}

Further, substituting by $\boldsymbol{u}(\boldsymbol{e}_q)$ from \eqref{eq:u_eq} and taking the expectation over the joint distribution of $\{s_{qnk}\}_{(n,k) \in \mathcal{R}_q}$ yields 
\begin{align}
    \mathbb{E} \left [ \left | L(\boldsymbol{e}_q) - L(\hat{\boldsymbol{e}}_q) \right | \right ] &\leq \mathbb{E} \left [ \| \nabla_{\boldsymbol{e}_q} L(\boldsymbol{e}_q) \|_2 \cdot \| \boldsymbol{e}_q - \hat{\boldsymbol{e}}_q \|_2 \right ] \nonumber \\ 
    &\ + \mathbb{E} \left [ \left \| \boldsymbol{u}(\boldsymbol{e}_q) \right \|_2 \cdot \| \boldsymbol{e}_q - \hat{\boldsymbol{e}}_q \|_2^2 \right ].
\end{align}

With non-negative expectations, the Cauchy-Schwarz inequality, i.e. $\mathbb{E} [A \cdot B] \leq \sqrt{\mathbb{E}[A^2] \cdot \mathbb{E}[B^2]}$, can be applied. 
Further, as $\nabla_{\boldsymbol{e}_q} L(\boldsymbol{e}_q)$ and $\boldsymbol{u}(\boldsymbol{e}_q)$ are independent of $s_{qnk}$, and with $\mathbb{E} \left [ \| \boldsymbol{e}_q - \hat{\boldsymbol{e}}_q \|_2^2 \right ] =  \mathbb{E} \left [ \| \boldsymbol{s}_q - \hat{\boldsymbol{s}}_q \|_2^2 \right ]$ from Proposition \ref{P1}, we obtain the statement of Proposition \ref{P2} as follows:
\begin{align}
    \mathbb{E} \left [ \left | L(\boldsymbol{e}_q) - L(\hat{\boldsymbol{e}}_q) \right | \right ] &\leq \| \nabla_{\boldsymbol{e}_q} L(\boldsymbol{e}_q) \|_2 \cdot \sqrt{\mathbb{E} \left [ \| \boldsymbol{s}_q - \hat{\boldsymbol{s}}_q \|_2^2 \right ]} \nonumber \\
    &\ + \| \boldsymbol{u}(\boldsymbol{e}_q) \|_2 \cdot \mathbb{E} \left [ \| \boldsymbol{s}_q - \hat{\boldsymbol{s}}_q \|_2^2 \right ].
\end{align}

\ifCLASSOPTIONcaptionsoff
  \newpage
\fi

\bibliographystyle{IEEEtran}
\bibliography{IEEEabrv,bibliography}

\vspace{-3em}

\begin{IEEEbiographynophoto}{Aladin Djuhera}
received his M.Sc. degree from the Technical University of Munich (TUM) in 2023, where he is currently pursuing his Ph.D. His research focuses on developing practical solutions for the scalable, efficient, and safe deployment of AI models, particularly in edge computing. Previously, he was with IBM Research, where he worked on federated learning, distributed inference, and AI workload orchestration for large language models.
\end{IEEEbiographynophoto}

\vspace{-3em}

\begin{IEEEbiographynophoto}{Vlad~C.~Andrei}
received his M.Sc. degree from the Technical University of Munich (TUM) in 2021 and is currently pursuing his Ph.D. His research interests include 6G, integrated sensing and communication, multi-agent robotic systems, digital twins, and their hardware implementation. He received the Best Paper and Best Student Demo Awards at the IEEE Symposium on Joint Communications and Sensing in 2023 and 2024, respectively.
\end{IEEEbiographynophoto}

\vspace{-3em}

\begin{IEEEbiographynophoto}{Xinyang~Li}
received his B.Sc. and M.Sc. degrees from Xidian University and the Technical University of Munich (TUM) in 2018 and 2022, respectively, with a focus on wireless communications and signal processing. He is currently pursuing his Ph.D. at TUM, researching system designs for integrated sensing and communication, machine learning, and information theory.
\end{IEEEbiographynophoto}

\vspace{-3em}

\begin{IEEEbiographynophoto}{Ullrich~J.~M{\"o}nich}
received his Dr.-Ing. degree from the Technical University of Munich (TUM) in 2011. From 2012 to 2015, he was a Post-Doctoral Fellow with the Massachusetts Institute of Technology. Since 2015, he has been a Senior Researcher and a Lecturer at TUM, where he leads research activities at the Advanced Communications and Embedded Security Laboratory (ACES Lab). His interests include wireless communications, physical-layer security, and integrated sensing and communication. He received the Rohde \& Schwarz Award for his dissertation in 2012.
\end{IEEEbiographynophoto}

\vspace{-3em}

\begin{IEEEbiographynophoto}{Holger~Boche}
is an IEEE Fellow and Full Professor at the Technical University of Munich (TUM), where he leads the Chair of Theoretical Information Technology. He received his Dr.-Ing. and Dr. rer. nat. degrees in electrical engineering and mathematics, respectively. His research interests include information theory, wireless communications, and quantum systems. He previously served as the Director of the Fraunhofer Institute for Telecommunications, Heinrich-Hertz-Institute (HHI), and is currently the Founding Director of the TUM Center for Quantum Engineering. He also co-leads the BMBF Research Hub 6G-life. He is a member of the German National Academy of Sciences Leopoldina and a recipient of the Gottfried Wilhelm Leibniz Prize. His work has been recognized with multiple IEEE Best Paper and Industry Awards, including the Vodafone Foundation Innovation Award.
\end{IEEEbiographynophoto}

\vspace{-3em}

\begin{IEEEbiographynophoto}{Walid~Saad}
is an IEEE Fellow and the Rolls-Royce Commonwealth Professor in Digital Twin Technology, Electrical and Computer Engineering at Virginia Tech, where he leads the Network intelligEnce, Wireless, and Security (NEWS) Lab. He received his Ph.D. degree from the University of Oslo, Norway in 2010. His research spans wireless networks (5G/6G), machine learning, game theory, quantum communications, and cyber-physical systems. He is a recipient of the NSF CAREER Award, the ONR Young Investigator Award, the IEEE Marconi Prize Award, and multiple IEEE Best Paper and Technical Achievement Awards. He is the Editor-in-Chief of the IEEE Transactions on Machine Learning in Communications and Networking and has been named a Clarivate Highly Cited Researcher annually since 2019.
\end{IEEEbiographynophoto}

\end{document}